\documentclass[journal]{IEEEtran}

\usepackage{color}
\usepackage{lineno}
\usepackage{cite}
\usepackage{verbatim}
\usepackage{amsmath,amssymb,amsfonts}
\usepackage{algorithmic}
\usepackage{graphicx}
\usepackage{textcomp}
\usepackage{xcolor}
\usepackage{subfigure}
\usepackage{setspace}
\usepackage{svg}
\usepackage[ruled]{algorithm2e}
\usepackage{booktabs}
\usepackage{url}
\usepackage{diagbox}
\usepackage{pifont}
\usepackage{xcolor}
\usepackage{multicol}
\usepackage{multirow}
\usepackage{tikz}
\usepackage[colorlinks=true,linkcolor=blue,citecolor=blue,urlcolor=blue]{hyperref}
\definecolor{lime}{HTML}{A6CE39}
\DeclareRobustCommand{\orcidicon}{%
    \begin{tikzpicture}
    \draw[lime, fill=lime] (0,0) 
    circle [radius=0.16] 
    node[white] {{\fontfamily{qag}\selectfont \tiny ID}};    \draw[white, fill=white] (-0.0625,0.095) 
    circle [radius=0.007];    \end{tikzpicture}
    \hspace{-2mm}}
\foreach \x in {A, ..., Z}{%
    \expandafter\xdef\csname orcid\x\endcsname{\noexpand\href{https://orcid.org/\csname orcidauthor\x\endcsname}{\noexpand\orcidicon}}
    }
\usepackage{nomencl}
\SetCommentSty{mycommfont}

\SetKwInput{KwInput}{Input}                
\SetKwInput{KwOutput}{Output}              

\begin{document}




\title{Through-the-Wall Radar Human Activity Recognition WITHOUT Using Neural Networks\\
\thanks{Manuscript received XXXXXXX XX, 2025; revised XXXXXXX XX, 2025; accepted XXXXXXX XX, 2025. Date of publication XXXXXXX XX, 2025; date of current version XXXXXXX XX, 2025.\par
My Bio: My name is Weicheng Gao. I'm a Ph.D. student from Beijing Institute of Technology. I’m majored and interested in mathematical and modeling theory research of signal processing, radar signal processing techniques, and AI for radar, apprenticed under professor Xiaopeng Yang. I’m currently dedicated in the field of Through-the-Wall Radar Human Activity Recognition. Looking forward to learning and collaborating with more like-minded teachers and mates. (e-mail: JoeyBG@126.com).\par
Digital Object Identifier 10.48550/arXiv.2506.XXXXX.\par}}

\author{Weicheng Gao\orcidA{},~\IEEEmembership{Graduate~Student~Member,~IEEE}
        }
        
\markboth{arXiv Preprint, June, 2025}%
{Shell \MakeLowercase{\textit{et al.}}: Bare Demo of IEEEtran.cls for IEEE Journals}

\maketitle

\begin{abstract}
After a few years of research in the field of through-the-wall radar (TWR) human activity recognition (HAR), I found that we seem to be stuck in the mindset of training on radar image data through neural network models. The earliest related works in this field based on template matching did not require a training process, and I believe they have never died. Because these methods possess a strong physical interpretability and are closer to the basis of theoretical signal processing research. In this paper, I would like to try to return to the original path by attempting to eschew neural networks to achieve the TWR HAR task and challenge to achieve intelligent recognition as neural network models. In detail, the range-time map and Doppler-time map of TWR are first generated. Then, the initial regions of the human target foreground and noise background on the maps are determined using corner detection method, and the micro-Doppler signature is segmented using the multiphase active contour model. The micro-Doppler segmentation feature is discretized into a two-dimensional point cloud. Finally, the topological similarity between the resulting point cloud and the point clouds of the template data is calculated using Mapper algorithm to obtain the recognition results. The effectiveness of the proposed method is demonstrated by numerical simulated and measured experiments. The open-source code of this work is released at: \href{https://github.com/JoeyBGOfficial/Through-the-Wall-Radar-Human-Activity-Recognition-Without-Using-Neural-Networks}{Github/JoeyBGOfficial/TWR-HAR-wo-NN-V1}.\par
\end{abstract}


\IEEEpeerreviewmaketitle

\begin{table}[!ht]
\begin{center}
\caption{Nomenclature.\label{Nomenclature}}
\vspace{-0.1cm}
\resizebox{0.48\textwidth}{!}{
\begin{tabular}{cc}
\hline\hline
\textbf{Abbreviation}             & \textbf{Full Name}     \\ 
\hline
ACM & Active Contour Model \\
AEN & Auto Encoder Network \\
BiGRU & Bidirectional Gated Recurrent Unit \\
CapsuleNet & Capsule Network \\
CNN & Convolutional Neural Network \\
ConvNeXt & Next Generation of Convolutional Network \\
CRF & Conditional Random Field \\
CWT & Continuous Wavelet Transform \\
DTM & Doppler-Time Map \\
DRLSE & Distance Regularized Level-Set Evolution \\
ECFRNet & Effective Corner Feature Representation Network \\
EMD & Empirical Modal Decomposition \\
FAST & Features from Accelerated Segment Test \\
GAC & Geodesic Active Contours \\
GCN & Graph Convolutional Neural Network \\
HAR & Human Activity Recognition \\
LBF & Local Binary Fitting \\
LFMCW & Linear Frequency Modulated Continuous Wave \\
MTI & Moving Target Indication \\
NLOS & Non-Line-of-Sight \\
ResNet & Residual Neural Network \\
ResNeXt & Next Generation of Residual Network \\
RPCA & Robust Principle Component Analysis \\
RTM & Range-Time Map \\
SIFT & Scale-Invariant Feature Transform \\
SISO & Single Input Single Output \\
SNR & Signal-to-Noise Ratio \\
STFT & Short-Time Fourier Transform \\
TCN & Temporal Convolutional Network \\
TWR & Through-the-Wall Radar \\
UCL & University College London \\
UWB & Ultra-Wideband \\
VGG & Visual Geometry Group \\
ViT & Vision Transformer \\
WSN & Wavelet Scattering Network \\
\hline\hline
\end{tabular}
}
\end{center}
\vspace{-0.0cm}
\end{table}\par

\section{Introduction}
With the advancement of wireless sensing, radar technology, deep learning and other technologies, TWR HAR has gradually become a research hotspot in the fields of intelligent security, rescue, and health monitoring \cite{Main1, Main2, Main3, Main4, Main5}. This technology can achieve contactless perception of human motion state in NLOS environments and has a wide range of application prospects \cite{Cuiguolong, Jintian, Wangjianqi, Dingyipeng, Yeshengbo}. Existing researches in this field have covered various aspects from signal preprocessing, feature extraction, deep network modeling to generalization ability enhancement, etc., and all of them have achieved certain stage-by-stage research results, demonstrating the active research value in this field \cite{Yangdegui, Jiayong, Qifugui}.\par
\begin{figure*}[!ht]
    \centering
    \includegraphics[width=\textwidth]{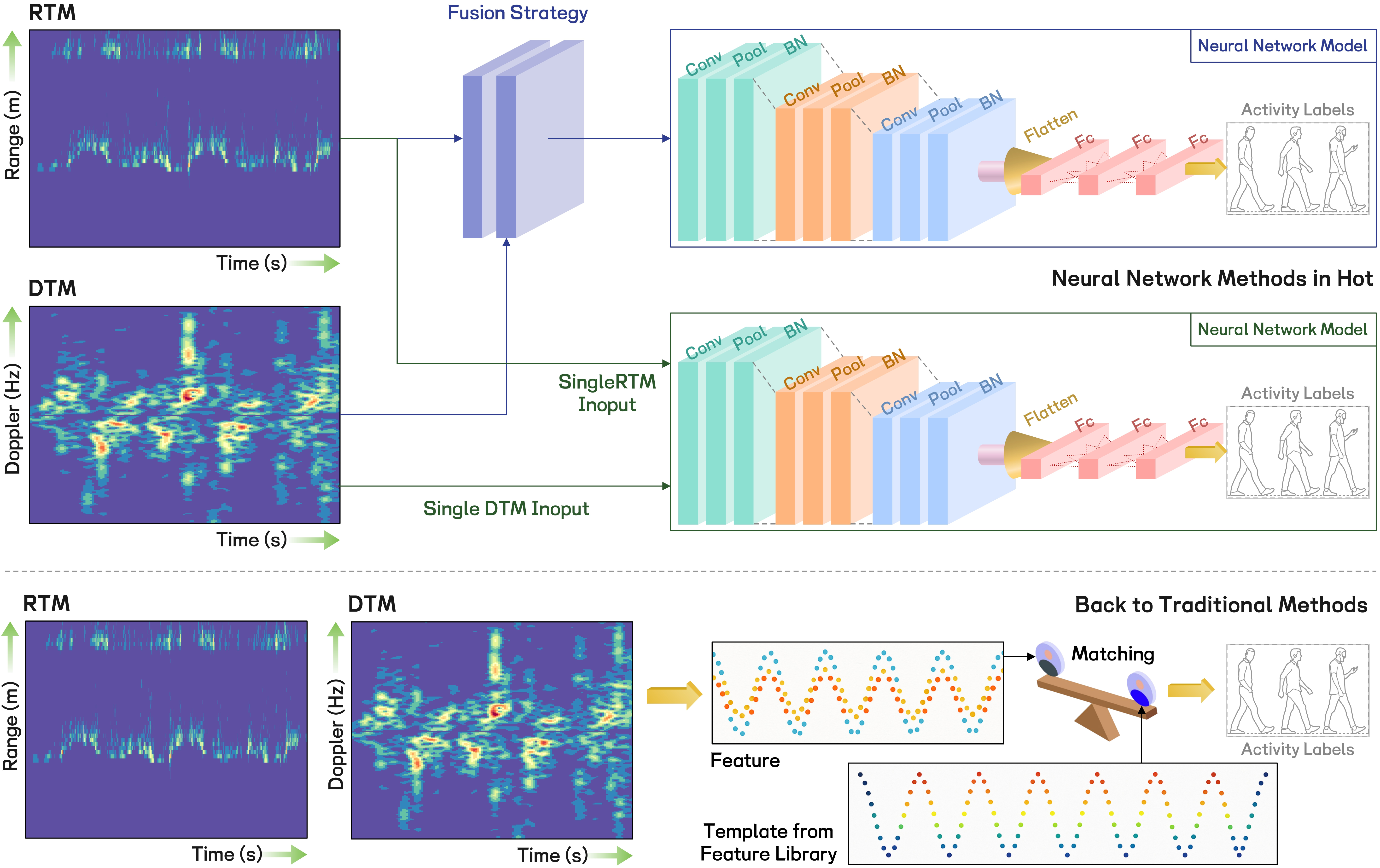}
    \caption{Current works in this field take neural network-based methods as the research hotspot. This work returns to rethink the value of traditional mindsets.}
    \label{Introduction}
    \vspace{-0.0cm}
\end{figure*}\par
In the past eight years, intelligent and accurate target recognition tasks have gradually become possible due to the rapid development in the fields of machine learning and deep learning. Related cutting-edge methods have also migrated to the field of TWR HAR and achieved a series of results \cite{Review1, Review2}. Cheng et al. proposed an end-to-end model based on range map sequences, which combined with a randomized tailoring training method to achieve $97.6\%$ recognition accuracy and output results in real time without waiting for the end of the activity \cite{Cheng}. Peng et al. enhanced the RTM by pixel-varying stripes to improve CNN model's recognition accuracy on nine types of activities \cite{Peng}. Yakoub et al. used CWT to construct time-frequency maps, which were processed by CNN to obtain a recognition rate as high as $99.59\%$ \cite{Yakoub}. Luo et al. proposed spectro-temporal network fusion of TCN and CNN to recognize $15$ classes of activities in kitchen scene with $99.64\%$ accuracy, verifying its generality and low latency advantage \cite{Luo}. Yang et al. combined AEN with sequential neural network module for real-time classification of time segments, achieving $93\%$ accuracy in only $20\%$ of the activity time \cite{Yang}. Cao et al. obtained $95.82\%$ recognition accuracy by principle component analysis denoising and EMD feature extraction, emphasizing on signal modeling for wall interference \cite{Cao}. Qi et al. analyzed micro-Doppler signature of fine-grained activities using an improved Hilbert-Huang transform to effectively deal with low SNR environments \cite{Qi}. Wang et al. integrated multiple features in time-frequency, range, and range-Doppler domains through GCN, and the recognition performance was better than the traditional method \cite{Wang}. Zhu et al. publicized the UWB TWR dataset of indoor human activities and proposed a CNN method with a testing accuracy higher than $99.7\%$, while calling for standardized dataset construction \cite{Zhu}. In addition to these, some cross-cutting works were also included: A cross-modal supervised learning approach to improve accuracy and robustness of human pose recognition was proposed by Xu et al. \cite{Xu}. Zhang et al. proposed a support vector machine algorithm to handle small high-dimensional samples and improve the efficiency of TWR HAR \cite{Zhang}.\par
For the TWR HAR task, the author's team has also been conducting research for some years and has achieved two sets of results. These included a series of one-stage algorithms that achieved data augmentation \cite{Gao1}, robustness \cite{Gao2} and computational speed improvement \cite{Gao3}, and a series of two-stage algorithms that achieved generalized recognition under different human targets \cite{Gao4, Gao5, Gao6}. All these existing works that have been hot in recent years coincidentally led to the same conclusion: HAR was an intelligent task, but due to the difficulty of modeling complex indoor human motion, the difficulty of wall clutter suppression, and the blurring and coupling of micro-Doppler signature with the difficulty of feature extraction, it was almost impossible to avoid training of deep learning algorithms in order to achieve accurate recognition \cite{Survey1, Survey2}. In fact, the earliest results in the field based on template matching required no training process. By setting the threshold value of a certain signal or radar image feature, it could directly determine whether a certain activity occurred and the category to which it belonged‌, or compared a certain feature in the real-time collected radar signal with the pre-defined template to identify activities with a high degree of similarity \cite{VCChen, EarlyRecognition}. These methods focused on manually designing rules and parameters rather than relying on data-driven model training. The limitations were that their rules are fixed, which made it difficult to handle complex or varied activities, and the recognition accuracy with scene adaptability tended to be ineffective. However, these methods were the most physically interpretable. Essentially, nowadays, neural networks provide a complex parameterized structure that can learn from large-scale datasets to physical mappings that we have difficulty parsimoniously expressing in signal processing. As effective as it is, its research is also moving further and further away from the original signal processing mindset.\par
\textbf{There are times when we need to return to our beginnings, even if the act of returning to our beginnings may not mean or contribute much. Maybe these methods aren't as bad as we thought.}\par
In this paper, a method to achieve TWR HAR without using neural network models is proposed. In terms of rigor, it should be emphasized that the proposed method does not have any neural network models, but it contains image feature extraction, functional analysis, clustering, and it also needs to generate a certain amount of data for point cloud matching. In essence, it is still a dismantling of the recognition logic of neural networks from a signal processing perspective. It is hoped that this will inform research in the field. Specific contributions of this paper are as follows:\par
\textbf{(1) TWR Human Echo Modeling:} In this paper, a detailed derivation of the human echo modeling for TWR is given. The method to generate RTM, DTM and suppress static clutter and noise is presented in detail. Both maps can be used for subsequent micro-Doppler signature extraction.\par
\textbf{(2) Refined Micro-Doppler Signature Extraction Based on ACM:} In this paper, the micro-Doppler signature foreground center estimation based on corner detection is first calculated, and the noise background center estimation is achieved at the same time. Refined micro-Doppler signature extraction using foreground and background centers as a starting points using Chan-Vese multiphase level set-based ACM is then proposed.\par
\textbf{(3) Point Cloud Topology Matching HAR:} In this paper, the extracted refined micro-Doppler signature are discretized into a point cloud using contour representation. The topological similarity between the point cloud and the point clouds generated from template data is used for directly mapping to obtain the activity label.\par
In addition, numerical simulated and measured experiments are carried out in this paper to demonstrate the effectiveness of the proposed method.\par
The rest of the paper is organized as follows: The TWR human echo model is first given in section II. The proposed ACM-based micro-Doppler signature extraction method and point cloud topology matching-based HAR method are then presented in section III. Numerical simulated and measured experiments are analyzed and discussed in section IV. Finally, the conclusion is given in section V.\par 
\begin{figure}[!ht]
    \centering
    \includegraphics[width=0.48\textwidth]{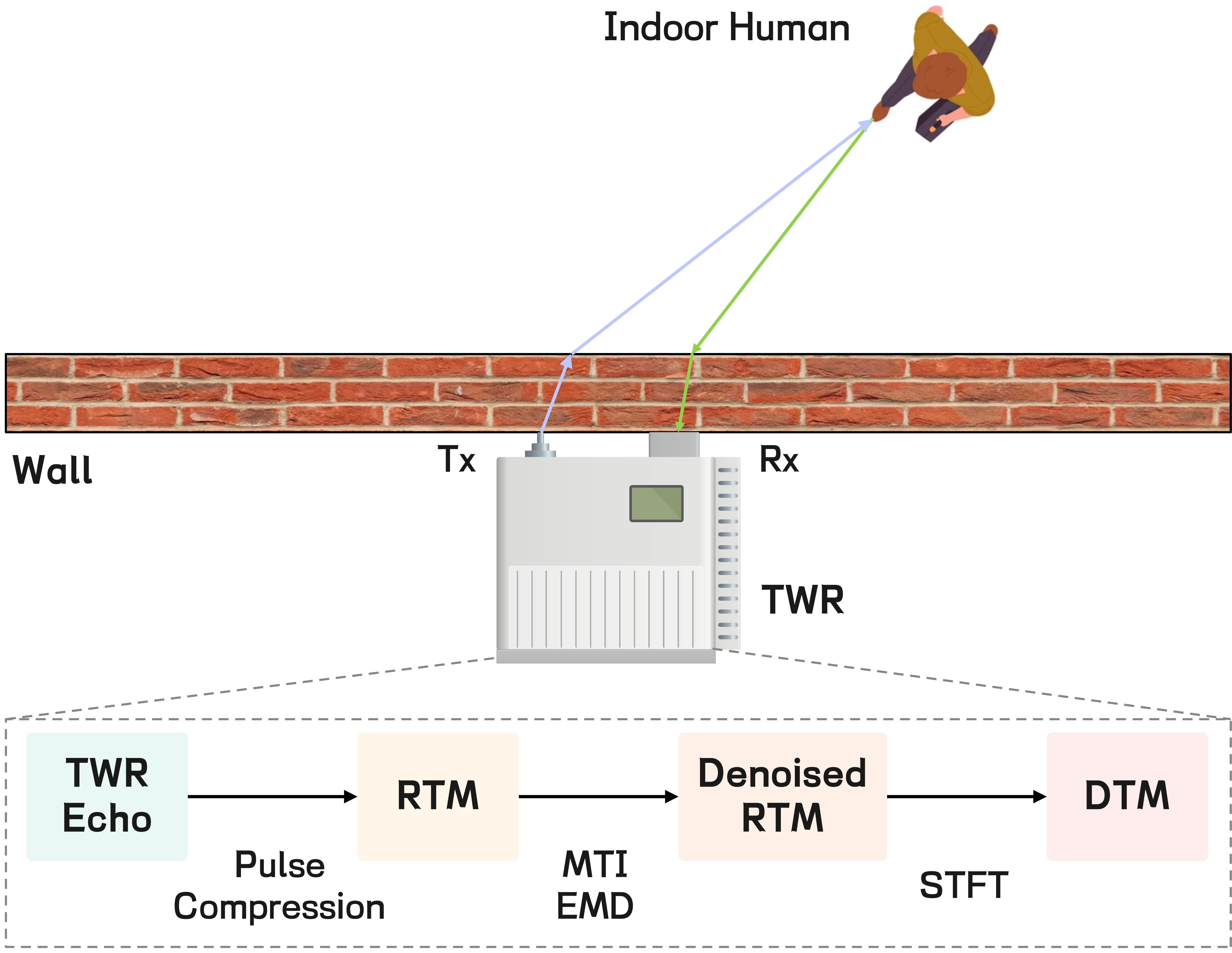}
    \caption{TWR human echo model and data processing.}
    \label{TWR_Echo_Model}
    \vspace{-0.2cm}
\end{figure}\par

\section{TWR Human Echo Model}
As shown in Fig. \ref{TWR_Echo_Model}, the UWB LFMCW is used for TWR transmission and reception \cite{TWR Echo Modeling}. The time-domain expression of the transmitted signal is:

\vspace{-0.2cm}
\begin{equation}
s(t) = \text{rect}\left(\frac{t}{T_p}\right) e^{j2\pi \left(f_c t + \frac{1}{2} \mu t^2 \right)},
\end{equation}
where $f_c$ is the carrier frequency, $T_p$ is the pulse width, $B$ is the bandwidth in $\mathrm{Hz}$ unit, $\mu=B/T_p$ is the slope of the frequency modulation. The definition of the rectangular function is:

\vspace{-0.2cm}
\begin{equation}
\text{rect}\left(\frac{t}{T_p}\right) =
    \begin{cases}
    1, & 0 \leq t \leq T_p \\
    0, & \text{otherwise}
    \end{cases}.
\end{equation}\par
After being scattered by the target, attenuated by the wall, and delayed in propagation, the echoes include reflections from multiple scattering centers in the human body, reflections from walls, and additive noise. Assuming that the head, the center of torso, left hand, right hand, left foot, and right foot of the human body correspond to $\mathrm{Hum}_i,i=1,2,\ldots,6$. The distance relative to the radar is $R_i(t),i=1,2,\ldots,6$, which varies over time due to human motion, and the backscattering cross-section is $\sigma_i,i=1,2,\ldots,6$, respectively. For each scattering center $i$, its echo signal is:

\vspace{-0.2cm}
\begin{equation}
\begin{aligned}
s_{r,i}(t) &= \sigma_i \alpha_w \text{rect}\left(\frac{t - \tau_i(t)}{T_p}\right)\\ &\cdot e^{j2\pi \left[f_c (t - \tau_i(t)) + \frac{1}{2} \mu (t - \tau_i(t))^2 \right]}
\end{aligned},
\end{equation}
where $\alpha_w\in (0,1]$ is the wall attenuation coefficient of signal amplitude, $\tau_i(t) = \frac{2R_i(t)}{c} + \tau_w$ is the time delay. The refraction delay introduced by the wall can be obtained through the fixed delay method \cite{Cuiguolong}:

\vspace{-0.2cm}
\begin{equation}
\tau_w = \frac{2d_w (\sqrt{\epsilon_r}-1)}{c},
\end{equation}
where $d_w,\varepsilon_r$ are the thickness and relative dielectric constant of the wall, respectively. The wall echo is considered as a reflection of a fixed scattering center. Assuming the backscattering cross-section is $\sigma_w$, the wall echo is:

\vspace{-0.2cm}
\begin{equation}
\begin{aligned}
s_{w}(t) &= \sigma_w \text{rect}\left(\frac{t - \tau_w}{T_p}\right) \\ & \cdot e^{j2\pi \left[f_c (t - \tau_w) + \frac{1}{2} \mu (t - \tau_w)^2 \right]}
\end{aligned}.
\end{equation}\par
Noise $\mathrm{no}(t)$ is usually additive Gaussian white noise, with zero mean and a certain variance. The total received signal is the superposition of human scattering center echo, wall echo, and noise:

\vspace{-0.4cm}
\begin{equation}
\begin{aligned}
s_r(t) &= \sum_{i=1}^{6} s_{r,i}(t) + s_{w}(t) + \mathrm{no}(t)\\
&= \sum_{i=1}^{6} \sigma_i \alpha_w \text{rect}\left(\frac{t - \tau_i(t)}{T_p}\right)e^{j2\pi \left[f_c (t - \tau_i(t)) + \frac{1}{2} \mu (t - \tau_i(t))^2 \right]} \\& +\sigma_w \text{rect}\left(\frac{t - \tau_w}{T_p}\right) e^{j2\pi \left[f_c (t - \tau_w) + \frac{1}{2} \mu (t - \tau_w)^2 \right]}+\mathrm{no}(t)
\end{aligned}.
\end{equation}\par
The pulse compression can be achived by matched filtering. The definition of the matched filter is:

\vspace{-0.2cm}
\begin{equation}
h(t) =s^{*}(-t)= \text{rect}\left(\frac{t}{T_p}\right) e^{j2\pi \left(f_c t - \frac{1}{2} \mu t^2 \right)}.
\end{equation}\par
The output of the matched filter is:

\vspace{-0.3cm}
\begin{equation}
\begin{aligned}
y(t)&= s_r(t)*h(t)\\
&=\int_{-\infty}^{\infty}  \left[ \sum_{i=1}^{6} \sigma_i \alpha_w \text{rect}\left(\frac{t - \tau_i(t)}{T_p}\right) \right. \\& \cdot e^{j2\pi \left[f_c (t - \tau_i(t)) + \frac{1}{2} \mu (t - \tau_i(t))^2 \right]} \\& +\sigma_w \text{rect}\left(\frac{t - \tau_w}{T_p}\right) e^{j2\pi \left[f_c (t - \tau_w) + \frac{1}{2} \mu (t - \tau_w)^2 \right]}\\& \left.+\mathrm{no}(t) \right] \cdot \left[\text{rect}\left(\frac{t-\tau}{T_p}\right) e^{j2\pi \left(f_c (t-\tau) - \frac{1}{2} \mu (t-\tau)^2 \right)}\right] \mathrm{d} \tau \\
&= \sum_{i=1}^{6} \sigma_i \alpha_w T_p \text{sinc}\left[B\left(t - \tau_i(t)\right)\right] e^{j2\pi f_c (t - \tau_i(t))} \\&+ \sigma_w T_p \text{sinc}\left[B\left(t - \tau_w\right)\right] e^{j2\pi f_c (t - \tau_w)} + \mathrm{no}'(t)
\end{aligned},
\end{equation}
holds true based on the Fourier transform of the $\mathrm{sinc}$ function, $\mathrm{sinc}(x)=\frac{\sin(\pi x)}{\pi x}$, and $\mathrm{no}^{\prime}(t)$ is the noise after filtering. The peak value of each $\mathrm{sinc}$ function corresponds to $\tau_i (t)$ and $\tau_w$, reflecting the distance of the scattering center and the wall from radar \cite{TWR Data Processing}.\par
Taking walking activity as an example, assuming that the human body moves uniformly in a straight line at a speed of $v$, the initial distance of the torso is $R_0$, $A_h,A_h^{\prime}$ are the swing amplitudes of the arms and legs, respectively, and the gait frequency is $f_h$, the head, both hands, and both feet are $\Delta R_1,\Delta R_3, \Delta R_5$ offset relative to the torso. The simplified human motion model can be expressed as:

\vspace{-0.4cm}
\begin{equation}
\begin{aligned}
R_1(t)&=R_0+vt+\Delta R_1\\
R_2(t)&=R_0+vt \\
R_3(t)&=R_0+vt+A_h\sin(2\pi f_h t)+\Delta R_3\\
R_4(t)&=R_0+vt+A_h\sin(2\pi f_h t-\pi)+\Delta R_3\\
R_5(t)&=R_0+vt+A^{\prime}_h\sin(2\pi f_h t)+\Delta R_5\\
R_6(t)&=R_0+vt+A^{\prime}_h\sin(2\pi f_h t-\pi)+\Delta R_5\\
\end{aligned}.
\end{equation}\par
The micro-Doppler signature is introduced by limb swinging, which affects the dynamic characteristics of the radar range profile \cite{RTM and DTM}.\par
Assuming that the radar transmits multiple pulses at pulse repetition interval $T_r$. Define the slow time index $t_m=mT_r$, where $m=0,1,2,\ldots,M-1$ and $M$ is the total number of pulses. For each slow time $t_m$, the signal $y(t,t_m)$ after pulse compression is recorded. Discretize the fast time as $t=nT_s$, $T_s$ is the sampling interval, $n=0,1,2,\ldots,N-1$, and the distance $R=\frac{c\tau}{2}$. The range unit is $R_n=\frac{cnT_s}{2}$. Thus the discrete form of the pulse compression echo is:

\vspace{-0.4cm}
\begin{equation}
\begin{aligned}
y(n,m) &= \sum_{i=1}^{6} \sigma_{i} \alpha_{w} T_p \, \mathrm{sinc} \left[ B \left( nT_{s} - \tau_{i}(t_{m}) \right) \right] \\&\cdot e^{j2\pi f_{c}(nT_{s} - \tau_{i}(t_{m}))} 
\\&+ \sigma_{w} T_p \, \mathrm{sinc} \left[ B \left( nT_{s} - \tau_{w} \right) \right] e^{j2\pi f_{c}(nT_{s} - \tau_{w})} 
\\&+ \mathrm{no}^{\prime}(n,m)
\end{aligned}.
\end{equation}\par
Static clutter components such as the wall can be removed from the echo using MTI filtering:

\vspace{-0.2cm}
\begin{equation}
y_\mathrm{MTI}(n,m) = y(n,m) - y(n,m-1).
\end{equation}\par
Absolute values are taken to obtain RTM for the MTI results $\mathrm{RTM}_\mathrm{MTI}(n,m)=|y_\mathrm{MTI}(n,m)|$. And after that, EMD is utilized to decompose the RTM and remove the noise \cite{EMD}:

\vspace{-0.4cm}
\begin{equation}
\begin{gathered}
\mathrm{RTM}_{\mathrm{MTI}}(n,m) = \sum_{k=1}^{K} \mathrm{IMF}_k(n,m)+r(n,m)\\
\mathrm{RTM}_{\mathrm{Denoised}}(n,m) = \sum_{k=k_0}^{K} \mathrm{IMF}_k(n,m)+r(n,m)
\end{gathered},
\end{equation}
where $\mathrm{IMF}_k(n,m)$ is the $k^\mathrm{th}$ intrinsic mode function, $K$ is the total number of modes, $k_0$ is the starting point of preserved low-frequency mode, $r(n,m)$ is the residual component.\par
The range cells of the denoised RTM are summed and the DTM is obtained by doing a STFT with the Hanning window $w(l)$ of length $L_\mathrm{Wind}$ and step $P_\mathrm{Wind}$:

\vspace{-0.2cm}
\begin{equation}
S(f_d, m^{\prime}) = \sum_{l=0}^{L-1} \left[\sum_{n=0}^{N-1}\mathrm{RTM}_{\mathrm{Denoised}}(n,m)\right] w(l) e^{-j2\pi f_d l T_r},
\end{equation}
where $f_d\in [-\frac{f_r}{2},\frac{f_r}{2}]$ is the Doppler frequency, $f_r=\frac{1}{T_r}$ is the pulse repetition frequency, $m^{\prime}=0,P_\mathrm{Wind},2P_\mathrm{Wind},\ldots$ is the center of the slow-time windows, $l$ is the in-window sampling index. Finally, the results are modeled to obtain $\mathrm{DTM}(f_d,m^{\prime})=|S(f_d, m^{\prime})|$.\par 
Both RTM and DTM can be used for subsequent micro-Doppler signature extraction and indoor HAR.\par

\section{Proposed Method}
In the proposed method, the micro-Doppler foreground and noise background centers using image corner detection are first estimated. Using the two centers as starting points, the micro-Doppler signature extraction is implemented based on ACM. Then, the extracted micro-Doppler signature is discretized into a point cloud based on contour representation, and the template data is topology-matched to achieve HAR.\par
\begin{figure*}
    \centering
    \includegraphics[width=\textwidth]{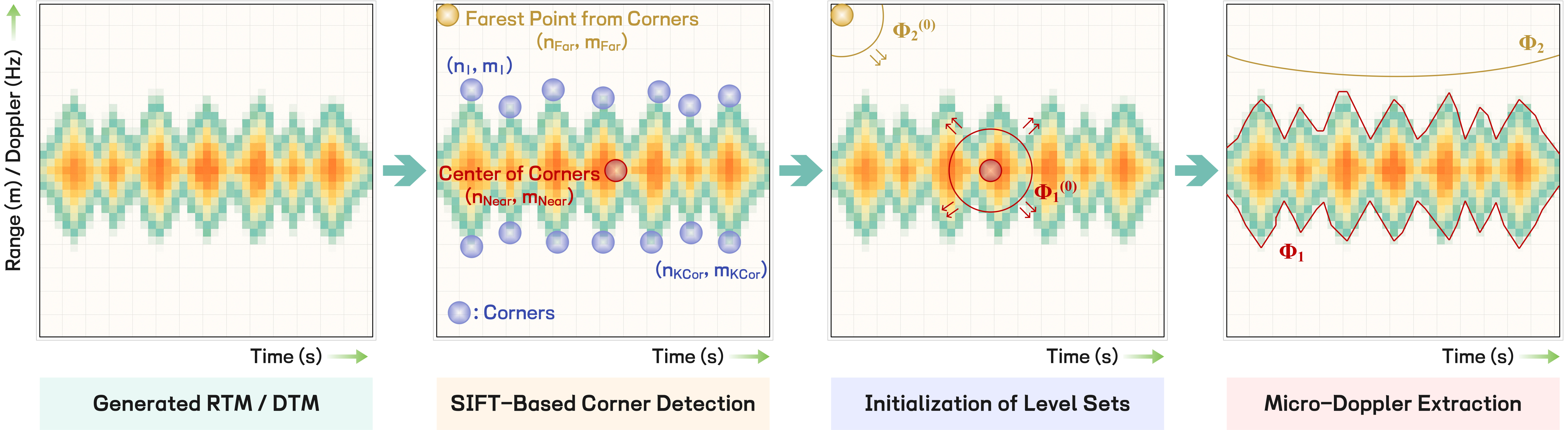}
    \caption{Schematic diagram of the proposed ACM-based micro-Doppler signature extraction method.}
    \label{ACM_Schematic}
    \vspace{-0.2cm}
\end{figure*}\par

\subsection{Micro-Doppler Signature Extraction Based on ACM}
Unify both input maps RTM and DTM to matrix variable $I(n,m)$, where $n=0,1,2,\ldots, N-1$ and $m=0,1,2,\ldots,M-1$. The $I(n,m)$ is first thresholded and truncated. Find the maximum pixel value by traversing each pixel of the image:

\vspace{-0.2cm}
\begin{equation}
\begin{gathered}
I_{\max} = \max_n \{\max_m \{I(n,m)\}\}\\
n\in [0,N-1] \cap \mathbb{Z}^{+}, m\in [0,M-1] \cap \mathbb{Z}^{+}
\end{gathered}.
\end{equation}\par
Define the threshold value $I_\mathrm{Threshold}$ as $\mathrm{Cut}_\mathrm{Threshold}$ times the maximum pixel value, and truncate the image based on this threshold value. Pixels smaller than the threshold are set to zero:

\vspace{-0.2cm}
\begin{equation}
I_\mathrm{Threshold} = \mathrm{Cut}_\mathrm{Threshold} \times I_{\max},
\end{equation}

\vspace{-0.2cm}
\begin{equation}
I'(n, m) = \begin{cases} 0 & \text{For} ~ I(n, m) < I_\mathrm{Threshold}  \\ 1 & \text{For} ~ I(n, m) \geq I_\mathrm{Threshold} \end{cases}.
\end{equation}\par
Next, the SIFT method is utilized to detect corner features on $I'(n,m)$. Micro-Doppler signature key points possessing scale and rotation invariance are detected by constructing an image pyramid \cite{Gao5, SIFT}. The multi-scale representation is generated by applying Gaussian blurring to the image at different scales:

\vspace{-0.3cm}
\begin{equation}
\mathrm{Gauss}(n, m, I_\sigma) = \frac{1}{2\pi I_\sigma^2} e^{-\frac{n^2 + m^2}{2I_\sigma^2}} \ast I'(n, m),
\end{equation}
where $\mathrm{Gauss}(n, m, I_\sigma)$ is the image after Gaussian blurring, $I_\sigma$ is the standard deviation that controls the degree of blurring, which divides the scale into multiple octaves with multiple sub-levels within each octave. Corners are detected by reduction of Gaussian blurred images at neighboring scales:

\vspace{-0.4cm}
\begin{equation}
\mathrm{DoG}(n, m, I_\sigma) = \mathrm{Gauss}(n, m, k I_\sigma) - \mathrm{Gauss}(n, m, I_\sigma),
\end{equation}
where $k=2^{1/\mathrm{oct}}$ is a constant, $\mathrm{oct}$ is the number of layers per octave. For each pixel $\mathrm{DoG}(n,m,I_\sigma)$, its $26$ neighbors at the current scale and adjacent $3\times 3 \times 3$ cubes are examined, and the pixel is marked as a key point if it is the maximum or minimum of these $26$ neighbors. If $|\mathrm{DoG}(n,m,I_\sigma)|$ is too small, eliminate the selected key point. Define Hessian matrix:

\vspace{-0.2cm}
\begin{equation}
\mathrm{Hes} = \left[\begin{array}{ll}
    \mathrm{DoG}_{nn} & \mathrm{DoG}_{nm}\\
    \mathrm{DoG}_{mn} & \mathrm{DoG}_{mm}
\end{array}\right],
\end{equation}
where $\mathrm{DoG}_{nn},\mathrm{DoG}_{nm},\mathrm{DoG}_{mn},\mathrm{DoG}_{mm}$ are four second-order derivatives of $\mathrm{DoG}(n,m,I_\sigma)$, respectively. Calculate the ratio of the two eigenvalues of the Hessian matrix and eliminate the keypoints if the ratio is too large. The final set of corners is obtained:

\vspace{-0.4cm}
\begin{equation}
\mathrm{Cor} = \{(n_1,m_1),(n_2,m_2),\ldots,(n_{K_\mathrm{Cor}},m_{K_\mathrm{Cor}})\},
\end{equation}
where $K_\mathrm{Cor}$ is the total number of corners. Calculate the coordinates of the center of gravity of all corners:

\vspace{-0.2cm}
\begin{equation}
n_{\text{Avg}} = \frac{1}{K_\mathrm{Cor}} \sum_{i=1}^{K_\mathrm{Cor}} n_i,~m_{\text{Avg}} = \frac{1}{K_\mathrm{Cor}} \sum_{i=1}^{K_\mathrm{Cor}} m_i. 
\end{equation}\par
Calculate the average Euclidean distance from each pixel to all corners in the image:

\vspace{-0.2cm}
\begin{equation}
\begin{gathered}
d_i(n,m)=\sqrt{(n-n_i)^2+(m-m_i)^2},\\
\text{Avg}_d(n, m) =\frac{1}{K_\mathrm{Cor}} \sum_{i=1}^{K_\mathrm{Cor}} \sqrt{(n - n_i)^2 + (m - m_i)^2}.
\end{gathered}
\end{equation}\par
Find the pixel with the largest average distance:

\vspace{-0.2cm}
\begin{equation}
(n_{\text{Far}}, m_{\text{Far}}) = \arg\max_{(n, m)} \text{Avg}_d (n, m).
\end{equation}\par
Calculate the Euclidean distance from each pixel to the center of gravity in the image:

\vspace{-0.2cm}
\begin{equation}
d(n, m) = \sqrt{(n - n_{\text{Avg}})^2 + (m - m_{\text{Avg}})^2}.
\end{equation}\par
Find the pixel with the smallest distance:

\vspace{-0.2cm}
\begin{equation}
(n_{\text{Near}}, m_{\text{Near}}) = \arg\min_{(n, m)} d(n, m).
\end{equation}\par
Both obtained coordinates $(n_\mathrm{Far},m_\mathrm{Far})$ and $(n_\mathrm{Near},m_\mathrm{Near})$ are recorded and will be used as the optimization starting points of subsequent ACM-based feature extraction \cite{ACM}.\par
The proposed ACM feature extraction method utilizes two level set functions $\phi_1,\phi_2$ to divide the image pixel space $\Omega$ into four parts \cite{Multiphase CV}, including:

\vspace{-0.2cm}
\begin{equation}
\begin{aligned}
& \Omega_{++}=\left\{x: \phi_1(x) \geq 0, \phi_2(x) \geq 0\right\} \\
& \Omega_{+-}=\left\{x: \phi_1(x) \geq 0, \phi_2(x)<0\right\} \\
& \Omega_{-+}=\left\{x: \phi_1(x)<0, \phi_2(x) \geq 0\right\} \\
& \Omega_{--}=\left\{x: \phi_1(x)<0, \phi_2(x)<0\right\}
\end{aligned},
\label{Regions Definition}
\end{equation}
where $x$ represents the image pixel in continuous form. Based on the classical Chan-Vese model, the continuous form of the dual level set four-phase segmentation energy function can be written as:

\vspace{-0.2cm}
\begin{equation}
\begin{aligned}
& E\left(\phi_1, \phi_2, c_{++}, c_{+-}, c_{-+}, c_{--}\right) \\ & = \mu_1 \int_{\Omega}\left|\nabla H\left(\phi_1\right)\right| \mathrm{d} x+\mu_2 \int_{\Omega}\left|\nabla H\left(\phi_2\right)\right| \mathrm{d} x \\
& +\lambda_{++} \int_{\Omega}\left|I(x)-c_{++}\right|^2 H\left(\phi_1\right) H\left(\phi_2\right) \mathrm{d} x \\
& +\lambda_{+-} \int_{\Omega}\left|I(x)-c_{+-}\right|^2 H\left(\phi_1\right)\left[1-H\left(\phi_2\right)\right] \mathrm{d} x \\
& +\lambda_{-+} \int_{\Omega}\left|I(x)-c_{-+}\right|^2\left[1-H\left(\phi_1\right)\right] H\left(\phi_2\right) \mathrm{d} x \\
& +\lambda_{--} \int_{\Omega}\left|I(x)-c_{--}\right|^2\left[1-H\left(\phi_1\right)\right]\left[1-H\left(\phi_2\right)\right] \mathrm{d} x
\end{aligned},
\label{Energy Function}
\end{equation}
where $H(\cdot)$ is the Heaviside function, which is generally achieved using its smooth approximation $H_\epsilon(s) \approx \frac{1}{2}\left[1+\frac{2}{\pi} \arctan \left(\frac{s}{\epsilon}\right)\right]$, and its derivative is the Dirac function $\delta_\epsilon(s)=H_\epsilon^{\prime}(s) \approx \frac{1}{\pi} \frac{\epsilon}{\epsilon^2+s^2}$, $\epsilon\in \mathbb{R^{+}}$ is a small value. $\mu_1,\mu_2$ control the strength of regularization on the boundary lengths of $\phi_1,\phi_2$, $\lambda_{++}, \lambda_{+-}, \lambda_{-+}, \lambda_{--}$ control the weights of the data fitting terms for four regions, $c_{++}, c_{+-}, c_{-+}, c_{--}$ are the constant average gray scale over four regions. The first two terms of this energy function represent the spatial rate of change of the phase field variables, which are typically used to describe the width and energy of the interface of both level sets. The latter four items control the uniformity of the four regions, resulting in smooth segmentation results.\par
Initially the recorded two coordinates are assigned to different sub-regions, in this paper two circles far away from each other are used to assign initial values to the level set:

\vspace{-0.2cm}
\begin{equation}
\begin{aligned}
\phi_1^{(0)}(n, m) & = \begin{cases}1, & \left(n-n_{\mathrm{Near}}\right)^2+\left(m-m_{\mathrm{Near}}\right)^2<\rho^2_1 \\
-1, & \text { Otherwise }\end{cases} \\
\phi_2^{(0)}(n, m) & = \begin{cases}1, & \left(n-n_{\mathrm{Far}}\right)^2+\left(m-m_{\mathrm{Far}}\right)^2<\rho^2_2 \\
-1, & \text { Otherwise }\end{cases}
\end{aligned},
\end{equation}
where $\rho_1,\rho_2$ are the initial small radius. With iterations, these two level sets evolve to give multiphase segmentation results. The segmentation energy minimization is used to estimate the level set functions $\phi_1,\phi_2$ and the variables $c_{++}, c_{+-}, c_{-+}, c_{--}$ by alternating iterations:\par
\vspace{0.4cm}
\textbf{(1) Freeze level sets to optimize average gray scale:}\par
When $\phi_1,\phi_2$ are fixed. Based on Eq. (\ref{Regions Definition}), at this point, the energy associated with $c_{++}, c_{+-}, c_{-+}, c_{--}$ is only the data fitting term:

\vspace{-0.2cm}
\begin{equation}
\begin{aligned}
E_{\text {data }}&=\lambda_{++} \int_{\Omega_{++}}\left|I-c_{++}\right|^2 \mathrm{d} x\\&+\lambda_{+-} \int_{\Omega_{+-}}\left|I-c_{+-}\right|^2 \mathrm{d} x\\&+\lambda_{-+} \int_{\Omega_{-+}}\left|I-c_{-+}\right|^2 \mathrm{d} x\\&+\lambda_{--} \int_{\Omega_{--}}\left|I-c_{--}\right|^2 \mathrm{d} x
\end{aligned}.
\end{equation}\par
Derive for $c_{++}, c_{+-}, c_{-+}, c_{--}$ one by one:

\vspace{-0.2cm}
\begin{equation}
\begin{gathered}
\frac{\partial}{\partial c_{++}}\left[\lambda_{++} \int_{\Omega_{++}}\left|I-c_{++}\right|^2 \mathrm{d} x\right]=0 \\
\frac{\partial}{\partial c_{+-}}\left[\lambda_{+-} \int_{\Omega_{+-}}\left|I-c_{+-}\right|^2 \mathrm{d} x\right]=0 \\
\frac{\partial}{\partial c_{-+}}\left[\lambda_{-+} \int_{\Omega_{-+}}\left|I-c_{-+}\right|^2 \mathrm{d} x\right]=0 \\
\frac{\partial}{\partial c_{--}}\left[\lambda_{--} \int_{\Omega_{--}}\left|I-c_{--}\right|^2 \mathrm{d} x\right]=0 \\
\end{gathered}.
\end{equation}\par
Thus:

\vspace{-0.2cm}
\begin{equation}
\begin{gathered}
c_{++}=\frac{\int_{\Omega_{++}} I(x) \mathrm{d} x}{\int_{\Omega_{++}} \mathrm{d} x}, ~ c_{+-}=\frac{\int_{\Omega_{+-}} I(x) \mathrm{d} x}{\int_{\Omega_{+-}} \mathrm{d} x},\\ c_{-+}=\frac{\int_{\Omega_{-+}} I(x) \mathrm{d} x}{\int_{\Omega_{-+}} \mathrm{d} x}, ~ c_{--}=\frac{\int_{\Omega_{--}} I(x) \mathrm{d} x}{\int_{\Omega_{--}} \mathrm{d} x}.
\end{gathered}
\end{equation}\par
\vspace{0.4cm}
\textbf{(2) Freeze average gray scale to optimize level sets:}\par
When $c_{++}, c_{+-}, c_{-+}, c_{--}$ are fixed. Known:

\vspace{-0.2cm}
\begin{equation}
\begin{gathered}
\left|\nabla H\left(\phi_i\right)\right| \approx \int \delta_\epsilon\left(\phi_i\right)\left|\nabla \phi_i\right| \mathrm{d} x\\
H\left(\phi_1\right) H\left(\phi_2\right) \approx H_\epsilon\left(\phi_1\right) H_\epsilon\left(\phi_2\right)
\end{gathered}.
\end{equation}\par
Based on Eq. (\ref{Energy Function}) and the Euler-Lagrange equation for functional optimization \cite{Euler-Lagrange}, assuming that:

\vspace{-0.2cm}
\begin{equation}
\begin{aligned}
E\left(\phi_1, \phi_2\right)&=F_1\left(\phi_1, \phi_2\right)+F_2\left(\phi_1, \phi_2\right)\\&+F_3\left(\phi_1, \phi_2\right)+F_4\left(\phi_1, \phi_2\right)\\&+\mu_1 \int \delta_\epsilon\left(\phi_1\right)\left|\nabla \phi_1\right| \mathrm{d} x\\&+\mu_2 \int \delta_\epsilon\left(\phi_2\right)\left|\nabla \phi_2\right| \mathrm{d} x
\end{aligned},
\end{equation}
where:

\vspace{-0.2cm}
\begin{equation}
\begin{aligned}
& F_1=\lambda_{++} \int\left|I-c_{++}\right|^2 H_\epsilon\left(\phi_1\right) H_\epsilon\left(\phi_2\right) \mathrm{d} x \\
& F_2=\lambda_{+-} \int\left|I-c_{+-}\right|^2 H_\epsilon\left(\phi_1\right)\left[1-H_\epsilon\left(\phi_2\right)\right] \mathrm{d} x \\
& F_3=\lambda_{-+} \int\left|I-c_{-+}\right|^2\left[1-H_\epsilon\left(\phi_1\right)\right] H_\epsilon\left(\phi_2\right) \mathrm{d} x \\
& F_4=\lambda_{--} \int\left|I-c_{--}\right|^2\left[1-H_\epsilon\left(\phi_1\right)\right]\left[1-H_\epsilon\left(\phi_2\right)\right] \mathrm{d} x
\end{aligned}.
\end{equation}\par
\begin{figure*}
    \centering
    \includegraphics[width=\textwidth]{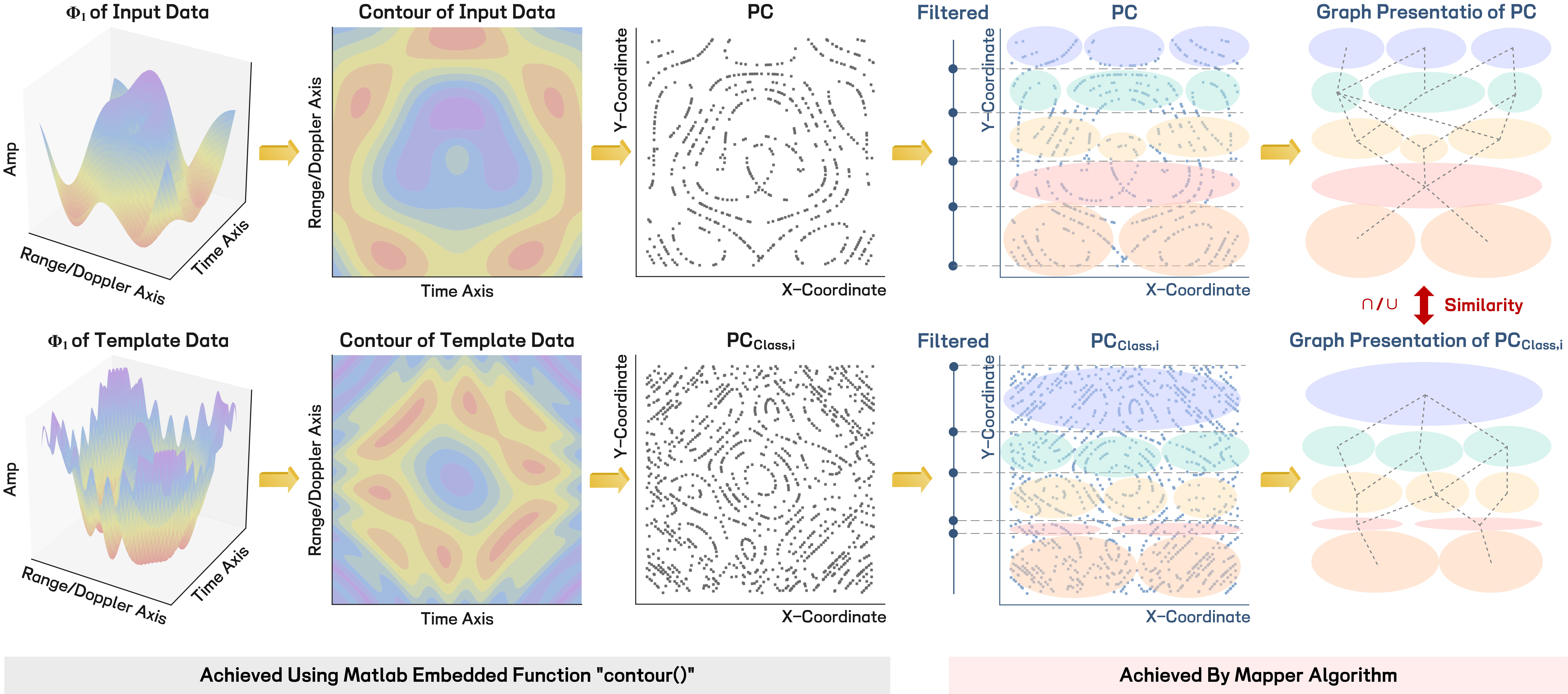}
    \caption{Schematic diagram of the proposed indoor HAR method based on point cloud topological structure similarity using Mapper algorithm.}
    \label{Pointcloud_Matching}
    \vspace{-0.0cm}
\end{figure*}\par
First analyze the variants of $\phi_1$. Variational fractions are obtained for regular terms:

\vspace{-0.2cm}
\begin{equation}
\frac{\delta}{\delta \phi_1}\left[\mu_1 \int \delta_\epsilon\left(\phi_1\right)\left|\nabla \phi_1\right| \mathrm{d} x\right]=\mu_1 \operatorname{div}\left(\delta_\epsilon\left(\phi_1\right) \frac{\nabla \phi_1}{\left|\nabla \phi_1\right|}\right).
\end{equation}\par
Variational fractions for the data items can be obtained by taking $F_1$ as an example. Because $H_\epsilon\left(\phi_2\right)$ can be seen as a constant for $\phi_1$:

\vspace{-0.2cm}
\begin{equation}
\begin{aligned}
\frac{\delta F_1}{\delta \phi_1}&=\lambda_{++} \int\left|I-c_{++}\right|^2 H_\epsilon\left(\phi_2\right) \delta_\epsilon\left(\phi_1\right) \mathrm{d} x\\&=\lambda_{++} H_\epsilon\left(\phi_2\right) \delta_\epsilon\left(\phi_1\right)\left|I-c_{++}\right|^2
\end{aligned}.
\end{equation}\par
Similarly:

\vspace{-0.2cm}
\begin{equation}
\begin{aligned}
\frac{\delta F_2}{\delta \phi_1}&=\lambda_{+-} (1-H_\epsilon\left(\phi_2\right)) \delta_\epsilon\left(\phi_1\right)\left|I-c_{+-}\right|^2\\
\frac{\delta F_3}{\delta \phi_1}&=-\lambda_{-+} H_\epsilon\left(\phi_2\right) \delta_\epsilon\left(\phi_1\right)\left|I-c_{-+}\right|^2\\
\frac{\delta F_2}{\delta \phi_1}&=-\lambda_{--} (1-H_\epsilon\left(\phi_2\right)) \delta_\epsilon\left(\phi_1\right)\left|I-c_{--}\right|^2\\
\end{aligned}.
\end{equation}\par
Thus:

\vspace{-0.2cm}
\begin{equation}
\begin{aligned}
\frac{\partial E}{\partial \phi_1}&=\mu_1 \operatorname{div}\left(\delta_\epsilon\left(\phi_1\right) \frac{\nabla \phi_1}{\left|\nabla \phi_1\right|}\right)\\&+\delta_\epsilon\left(\phi_1\right)\cdot \left[\lambda_{++} H_\epsilon\left(\phi_2\right)\left|I-c_{++}\right|^2\right.\\&+\lambda_{+-}\left(1-H_\epsilon\left(\phi_2\right)\right)\left|I-c_{+-}\right|^2\\&-\lambda_{-+} H_\epsilon\left(\phi_2\right)\left|I-c_{-+}\right|^2\\&\left.-\lambda_{--}\left(1-H_\epsilon\left(\phi_2\right)\right)\left|I-c_{--}\right|^2\right]
\end{aligned}.
\end{equation}\par
Define the artificial time step $t_\mathrm{Step}$ by $\frac{\partial \phi_1}{\partial t_\mathrm{Step}}=-\frac{\partial E}{\partial \phi_1}$ to get the gradient descent update formula \cite{GD}.\par
Then analyze the variants of $\phi_2$. Consistent with the derivation process above, it can be obtained:

\vspace{-0.2cm}
\begin{equation}
\begin{aligned}
\frac{\partial E}{\partial \phi_2}&=\mu_2 \operatorname{div}\left(\delta_\epsilon\left(\phi_2\right) \frac{\nabla \phi_2}{\left|\nabla \phi_2\right|}\right)\\&+\delta_\epsilon\left(\phi_2\right)\cdot \left[\lambda_{++} H_\epsilon\left(\phi_1\right)\left|I-c_{++}\right|^2\right.\\&+\lambda_{+-}\left(1-H_\epsilon\left(\phi_1\right)\right)\left|I-c_{+-}\right|^2\\&-\lambda_{-+} H_\epsilon\left(\phi_1\right)\left|I-c_{-+}\right|^2\\&\left.-\lambda_{--}\left(1-H_\epsilon\left(\phi_1\right)\right)\left|I-c_{--}\right|^2\right]    
\end{aligned}.
\end{equation}\par
Use the same predefined time step $t_\mathrm{Step}$ by $\frac{\partial \phi_2}{\partial t_\mathrm{Step}}=-\frac{\partial E}{\partial \phi_2}$ to get another gradient descent update formula.\par
To facilitate the numerical solution of the gradient descent, some of the operators need to be discretized:

\vspace{-0.2cm}
\begin{equation}
\begin{aligned}
\nabla \phi \approx \frac{1}{2}\left[\begin{array}{l}
    \phi(n+1, m)-\phi(n-1, m)\\\phi(n, m+1)-\phi(n, m-1)
\end{array} \right] \\   
\end{aligned},
\end{equation}

\vspace{-0.2cm}
\begin{equation}
\begin{aligned}
\operatorname{div}\left(\frac{\nabla \phi}{|\nabla \phi|}\right) & \approx \frac{\partial}{\partial n}\left(\frac{\phi_{n+1, m}-\phi_{n, m}}{\left|\nabla \phi_{n, m}\right|}\right)\\&+\frac{\partial}{\partial m}\left(\frac{\phi_{n, m+1}-\phi_{n, m}}{\left|\nabla \phi_{n, m}\right|}\right)
\end{aligned}.
\end{equation}\par
Algorithm \ref{ACM Numerical Solution} gives the detailed procedure for the numerical solution. For the results of two level set functions $\phi_1,\phi_2$ and four segmentation regions, $\Omega_{+-}$ is considered as the extracted micro-Doppler signature region.\par
\begin{algorithm*}[!ht]
\DontPrintSemicolon
  \KwInput{Image $I(n,m)$ and two recorded coordinates $(n_\mathrm{Near}, m_\mathrm{Near})$ and $(n_\mathrm{Far}, m_\mathrm{Far})$.} 
  \KwOutput{Results of two level set functions $\phi_1,\phi_2$.}
  Initializing $\mu_1, \mu_2, \lambda_{++}, \lambda_{+--}, \lambda_{-+}, \lambda_{--},\epsilon,t_\mathrm{Step}$; $\mathrm{Stop}_\mathrm{Threshold}$ for the end of solution; and two level set functions:\;
  \vspace{-0.1cm}
    $$\phi_1^{(0)} = \begin{cases}1, & \left(n-n_{\mathrm{Near}}\right)^2+\left(m-m_{\mathrm{Near}}\right)^2<\rho^2_1 \\
    -1, & \text { Otherwise }\end{cases},\quad
    \phi_2^{(0)} = \begin{cases}1, & \left(n-n_{\mathrm{Far}}\right)^2+\left(m-m_{\mathrm{Far}}\right)^2<\rho^2_2 \\
    -1, & \text { Otherwise }\end{cases};$$\;
  \vspace{-0.4cm}
  \While{$\mathrm{TRUE}$}{
    Updating the average value of four regions $c^{(k+1)}_{++},c^{(k+1)}_{+-},c^{(k+1)}_{-+},c^{(k+1)}_{--}$:
    $$\begin{gathered}
        c_{++}^{(k+1)}=\frac{\sum_{(n, m)} I(n, m) H_\epsilon\left(\phi_1^{(k)}(n,m)\right) H_\epsilon\left(\phi_2^{(k)}(n,m)\right)}{\sum_{(n, m)} H_\epsilon\left(\phi_1^{(k)}(n,m)\right) H_\epsilon\left(\phi_2^{(k)}(n,m)\right)}\\
        c_{+-}^{(k+1)}=\frac{\sum_{(n, m)} I(n, m) H_\epsilon\left(\phi_1^{(k)}(n,m)\right) \left(1-H_\epsilon\left(\phi_2^{(k)}(n,m)\right)\right)}{\sum_{(n, m)} H_\epsilon\left(\phi_1^{(k)}(n,m)\right) \left(1-H_\epsilon\left(\phi_2^{(k)}(n,m)\right)\right)}\\
        c_{-+}^{(k+1)}=\frac{\sum_{(n, m)} I(n, m) \left(1-H_\epsilon\left(\phi_1^{(k)}(n,m)\right)\right) H_\epsilon\left(\phi_2^{(k)}(n,m)\right)}{\sum_{(n, m)} \left(1-H_\epsilon\left(\phi_1^{(k)}(n,m)\right)\right) H_\epsilon\left(\phi_2^{(k)}(n,m)\right)}\\
        c_{--}^{(k+1)}=\frac{\sum_{(n, m)} I(n, m) \left(1-H_\epsilon\left(\phi_1^{(k)}(n,m)\right)\right) \left(1-H_\epsilon\left(\phi_2^{(k)}(n,m)\right)\right)}{\sum_{(n, m)} \left(1-H_\epsilon\left(\phi_1^{(k)}(n,m)\right)\right) \left(1-H_\epsilon\left(\phi_2^{(k)}(n,m)\right)\right)}\\
    \end{gathered};$$\;
    \vspace{-0.4cm}
    Updating two level set functions:
    $$\begin{gathered}
        \frac{\partial E}{\partial \phi_1}=\mu_1 \operatorname{div}\left(\delta_\epsilon\left(\phi_1\right) \frac{\nabla \phi_1}{\left|\nabla \phi_1\right|}\right)+\delta_\epsilon\left(\phi_1\right)\left[\lambda_{++} H_\epsilon\left(\phi_2\right)\left|I-c_{++}\right|^2+\lambda_{+-}\left(1-H_\epsilon\left(\phi_2\right)\right)\left|I-c_{+-}\right|^2\right. \\
        \left.-\lambda_{-+} H_\epsilon\left(\phi_2\right)\left|I-c_{-+}\right|^2-\lambda_{--}\left(1-H_\epsilon\left(\phi_2\right)\right)\left|I-c_{--}\right|^2\right]\\
        \frac{\partial E}{\partial \phi_2}=\mu_2 \operatorname{div}\left(\delta_\epsilon\left(\phi_2\right) \frac{\nabla \phi_2}{\left|\nabla \phi_2\right|}\right)+\delta_\epsilon\left(\phi_2\right)\left[\lambda_{++} H_\epsilon\left(\phi_1\right)\left|I-c_{++}\right|^2-\lambda_{+-} H_\epsilon\left(\phi_1\right)\left|I-c_{+-}\right|^2\right. \\
        \left.+\lambda_{-+}\left(1-H_\epsilon\left(\phi_1\right)\right)\left|I-c_{-+}\right|^2-\lambda_{--}\left(1-H_\epsilon\left(\phi_1\right)\right)\left|I-c_{--}\right|^2\right]\\
        \phi_1^{(k+1)} =\phi_1^{(k)}-\Delta t_\mathrm{Step} \cdot \frac{\partial E}{\partial \phi_1}\left(\phi_1^{(k)}, \phi_2^{(k)}, c^{(k+1)}_{++},c^{(k+1)}_{+-},c^{(k+1)}_{-+},c^{(k+1)}_{--}\right) \\
        \phi_2^{(k+1)} =\phi_2^{(k)}-\Delta t_\mathrm{Step} \cdot \frac{\partial E}{\partial \phi_2}\left(\phi_1^{(k)}, \phi_2^{(k)}, c^{(k+1)}_{++},c^{(k+1)}_{+-},c^{(k+1)}_{-+},c^{(k+1)}_{--}\right)
    \end{gathered};$$\;
    \vspace{-0.3cm}
    \If{$\max \left\{\left\|\phi_1^{(k+1)}-\phi_1^{(k)}\right\|,\left\|\phi_2^{(k+1)}-\phi_2^{(k)}\right\|\right\}<\mathrm{Stop}_\mathrm{Threshold}$ \rm{or the maximum iteration is reached}}{Break;\;}
    }
\caption{Numerical Solution for ACM-based Micro-Doppler Signature Extraction}
\label{ACM Numerical Solution}
\end{algorithm*}\par

\subsection{Indoor HAR Based on Point Cloud Matching}
In this paper, a point cloud topology similarity estimation based on Mapper's algorithm is proposed for indoor HAR \cite{Mapper}. The proposed method matches the input point cloud with the template point cloud to achieve classification \cite{Pointcloud Matching}.\par
The contour features of $\phi_1$ are first solved and the contours are discretized into a two-dimensional point cloud. Define $\mathrm{contour}()$ as the contour feature generation function implemented in MATLAB that $\mathrm{PC} = \mathrm{contour}(\phi_1) = \{p_1,p_2,\ldots,p_k,...p_{N_\mathrm{PC}}\}$, where $\mathrm{PC}$ is the contour point cloud, $p_k=(x_k,y_k),k=1,2,\ldots,N_\mathrm{PC}$ are the points, $x_k,y_k$ are horizontal and vertical coordinates, respectively.\par
Define the following linear constant mapping as a filter function of Mapper preprocessing:

\vspace{-0.2cm}
\begin{equation}
\mathrm{Filt}(p_k)=p_k,~k=1,2,\ldots,N_\mathrm{PC}.
\end{equation}\par
For a total of $\mathrm{Class}=1,2,\ldots,\mathrm{Cla}$ classes of activities, a fixed number of $i=1,2,\ldots,\mathrm{Cla}_\mathrm{Num}$ data is taken for each class, and both ACM-Based micro-Doppler signature extraction, point cloud generation method with the same hyperparameter settings are used to obtain templates $\mathrm{PC}_{\mathrm{Class},i}$. Calculate the minimum and maximum values of the input point cloud $\mathrm{PC}$ and the template $\mathrm{PC}_{\mathrm{Class},i}$:

\vspace{-0.4cm}
\begin{equation}
\begin{gathered}
\min _{x}=\min \left(\min (\mathrm{PC}[0,:]), \min \left(\mathrm{PC}_{\text {Class }, i}[0,:]\right)\right)
\\ \max _{x}=\max \left(\max (\mathrm{PC}[0,:]), \max \left(\mathrm{PC}_{\text {Class }, i}[0,:]\right)\right) 
\\
\min _{y}=\min \left(\min (\mathrm{PC}[1,:]), \min \left(\mathrm{PC}_{\text {Class }, i}[1,:]\right)\right)
\\ \max _{y}=\max \left(\max (\mathrm{PC}[1,:]), \max \left(\mathrm{PC}_{\text {Class }, i}[1,:]\right)\right)
\end{gathered},
\end{equation}
which coverages the range of $[\min_x,\max_x]\times [\min_y,\max_y]$. Divide the $x$ direction into $n_x$ intervals with the step size of $\mathrm{step}=(\max_x-\min_x)/(n_x-1)$, and divide the $y$ direction into $n_y$ intervals with the step size of $\mathrm{step}=(\max_y-\min_y)/(n_y-1)$. Each rectangular grid is sized as $s_{x}=\mathrm{step}_{x} \cdot \mathrm{of}$ and $s_{y}=\mathrm{step}_{y} \cdot \mathrm{of}$, where $\mathrm{of}>1$ ensures the overlapping of grid exists. Defining coverage sets $\mathrm{Cov}_{i,j}$ with the center of $\left(\min _{x}+i \cdot \operatorname{step}_{x}, \min _{y}+j \cdot \operatorname{step}_{y}\right)$ and range:

\vspace{-0.4cm}
\begin{equation}
\begin{aligned}
U_{i, j}&=\left[\min _{x}+i \cdot \operatorname{step}_{x}-s_{x} / 2, \min _{x}+i \cdot \operatorname{step}_{x}+s_{x} / 2\right] \\& \times\left[\min _{y}+j \cdot \operatorname{step}_{y}-s_{y} / 2, \min _{y}+j \cdot \operatorname{step}_{y}+s_{y} / 2\right]
\end{aligned},
\end{equation}
where $i=0,1,\ldots,n_x-1$ and $j=0,1,\ldots,n_y-1$.\par
Mapper algorithm clusters the points in $\mathrm{Cov}_{i,j} \cap \mathrm{PC}$ and $\mathrm{Cov}_{i,j} \cap \mathrm{PC}_{\mathrm{Class},i}$ and adds edges based on the intersection between clusters. Define horizontal edges $(i,j) \rightarrow (i+1,j)$, where $i=0,1,\ldots,n_x-2$ and $j=0,1,\ldots,n_y-1$. Define vertical edges $(i,j) \rightarrow (i,j+1)$, where $i=0,1,\ldots,n_x-1$ and $j=0,1,\ldots,n_y-2$. For horizontal edges, the overlap region is calculated as:

\vspace{-0.3cm}
\begin{equation}
\begin{aligned}
O_{(i, j),(i+1, j)}&=\Big[\min _{x}+(i+1) \cdot \operatorname{step}_{x}-s_{x} / 2, \\&\quad~ \min _{x} + i \cdot \operatorname{step}_{x}+s_{x} / 2\Big] \\&\times\Big[\min _{y}+j \cdot \operatorname{step}_{y}-s_{y} / 2, \\&\quad~ \min _{y}+j \cdot \operatorname{step}_{y}+s_{y} / 2\Big]
\end{aligned}.
\end{equation}\par
For vertical edges, the overlap region is calculated as:

\vspace{-0.3cm}
\begin{equation}
\begin{aligned}
O_{(i, j),(i+1, j)}&=\Big[\min _{x}+i \cdot \operatorname{step}_{x}-s_{x} / 2, \\&\quad~ \min _{x} + i \cdot \operatorname{step}_{x}+s_{x} / 2\Big] \\&\times\Big[\min _{y}+(j+1) \cdot \operatorname{step}_{y}-s_{y} / 2, \\&\quad~ \min _{y}+j \cdot \operatorname{step}_{y}+s_{y} / 2\Big]
\end{aligned}.
\end{equation}\par
For point cloud $\mathrm{PC}$ and the template $\mathrm{PC}_{\mathrm{Class},i}$, the set of edges is defined as:

\vspace{-0.1cm}
\begin{equation}
\begin{gathered}
\mathrm{Edge}_\mathrm{PC}=\left\{\mathrm{PC} \cap O_{e} \neq \emptyset\right\}\\
\mathrm{Edge}_{\mathrm{PC}_{\mathrm{Class},i}}=\left\{\mathrm{PC}_{\mathrm{Class},i} \cap O_{e} \neq \emptyset\right\}\\
\mathrm{For} ~e \in \mathrm{Horizontal/Vertical~Edges}
\end{gathered}.
\end{equation}\par
The topological similarity of the two point clouds is quantified using the Jaccard similarity:

\vspace{-0.3cm}
\begin{equation}
\mathrm{similarity}_{\mathrm{Class},i}=\frac{\left|\mathrm{Edge}_{\mathrm{PC}} \cap \mathrm{Edge}_{\mathrm{PC}_{\text {Class }, i}}\right|}{\left|\mathrm{Edge}_{\mathrm{PC}} \cup \mathrm{Edge}_{\mathrm{PC}_{\text {Class }, i}}\right|},
\end{equation}
denotes the ratio of the number of edges shared by two graphs to the size of the concatenation of the sets of edges of the two graphs, with the range of $[0,1]$. A larger value indicates a more similar topology. Finally, the category that sums up the maximum similarity over all the data is found to be the desired activity recognition result \cite{Classification}:

\vspace{-0.2cm}
\begin{equation}
\underset{\mathrm{Class}}{\arg \max} \sum_{i=1}^{\mathrm{Cla}_\mathrm{Num}}  \mathrm{similarity}_{\mathrm{Class},i}.  
\end{equation}\par
\begin{table}[!ht]
\begin{center}
\caption{Parameter and Scene Settings$^{*}$.\label{Parameter and Scene Settings}}
\resizebox{0.48\textwidth}{!}{
\begin{tabular}{cc}
\hline\hline
\textbf{Parameters}             & \textbf{Value}     \\ 
\hline
Antenna Transceiver Spacing     &  $0.15~m$ (SISO Mode)  \\
Waveform                        &  LFMCW           \\
Antenna Height to Ground & $1.5~m$      \\ 
Center Frequency          & $1.5 \mathrm{~GHz}$   \\
Band Width & $2.0 \mathrm{~GHz}$  \\
Fast-Time Sampling Points$^{1}$ & $1024$                \\
Slow-Time Sampling Points$^{1}$ & $256/s$             \\
Sampling Period$^{1}$ & $4~s$                 \\
Wall Thickness & $0.12~m$  \\
Wall Relative Dielectric Constant & $6$ (Estimated) \\
Human Motion Range from Radar & $1 \sim 4 ~m$     \\
Number of Activities ($\mathrm{Cla}$) & $12$ \\
Template Dataset Size$^{2}$ ($\mathrm{Cla}_\mathrm{Num}$) &  $20$ Per Activity    \\
Validation Dataset Size$^{3}$     & $800$     \\
\hline\hline
\end{tabular}
}
\end{center}
\footnotesize $^{*}$ Simulations and measurements are conducted under the same parameters.\\
\footnotesize $^{1}$ The total number of points in both fast time and slow time is $1024$, making the echo a square matrix. This ensures that the resize scale for both fast time and slow time dimensions is consistent in image processing.\\
\footnotesize $^{2}$ A total of $4000$ sets are collected. However, only $20$ sets are extracted per activity for template matching.\\
\footnotesize $^{3}$ $140$ sets for empty scene. The remaining $11$ activities each contain $60$ sets.\\
\vspace{-0.0cm}
\end{table}\par
\begin{table}[!ht]
\begin{center}
\caption{Hyperparameter Settings$^{*}$.\label{Hyperparameter Settings}}
\resizebox{0.48\textwidth}{!}{
\begin{tabular}{cc}
\hline\hline
\textbf{Hyperparameters}             & \textbf{Value}     \\ 
\hline
\multicolumn{2}{c}{\textbf{Parameters of the Proposed Method}}\\
\hline
$k_0$ & $3$ \\
$L_\mathrm{Wind}$ & $0.5~s$\\
$P_\mathrm{Wind}$ & $0.05~s$\\
$\mathrm{Cut}_\mathrm{Threshold}$ & $0.3$\\
$I_\sigma$ \cite{Gao5} & $1.6$ \\
$\mathrm{oct}$ \cite{Gao5} & $3$ \\
$K_\mathrm{Cor}$ \cite{Gao4,Gao5} & $30$ \\
$\lambda_{++}, \lambda_{+--}, \lambda_{-+}, \lambda_{--}$     &  $1$  \\
$\mu_1, \mu_2$     &  $0.5$  \\
$t_\mathrm{Step}$ & $0.1$\\
$\epsilon$ & $1$\\
\multirow{2}{*}{$\rho_1,\rho_2$} & Simulated RTM/DTM: $64$\\
& Measured RTM/DTM: $32$\\
\multirow{4}{*}{Maximum Iteration of Algorithm \ref{ACM Numerical Solution}} & Simulated RTM: $20$\\
& Simulated DTM: $20$\\
& Measured RTM: $30$\\
& Measured DTM: $50$\\
Evolution Steps of Level Sets & $70$\\
$n_x,n_y$ & $100$ \\
$\mathrm{of}$ & $1.5$ \\
\hline 
\multicolumn{2}{c}{\textbf{Hardware and Software Conditions}}\\
\hline
Execution CPU Environment & Intel Core i9-10850K \\
Execution GPU Environment & NVIDIA RTX 3060 OC \\
Execution Software & MATLAB R2024b \\
\hline\hline
\end{tabular}
}
\end{center}
\footnotesize $^{*}$ Hyperparameters are chosen at the input image scale of $256\times 256$. It is recommended to dynamically adjust the hyper-parameter settings according to different data features, input image scales, and hardware resources.\\
\vspace{-0.4cm}
\end{table}\par
Although the proposed method does not use neural networks for the whole process, it requires a multi-step optimization process and a certain amount of data for template matching. Essentially it physically dismantles a portion of the neural network implementation. Theoretically, the proposed method is definitely not comparable to the accuracy of neural networks, but it can provide a reference for trying out the idea.\par
\begin{figure*}[!ht]
    \centering
    \includegraphics[width=\textwidth]{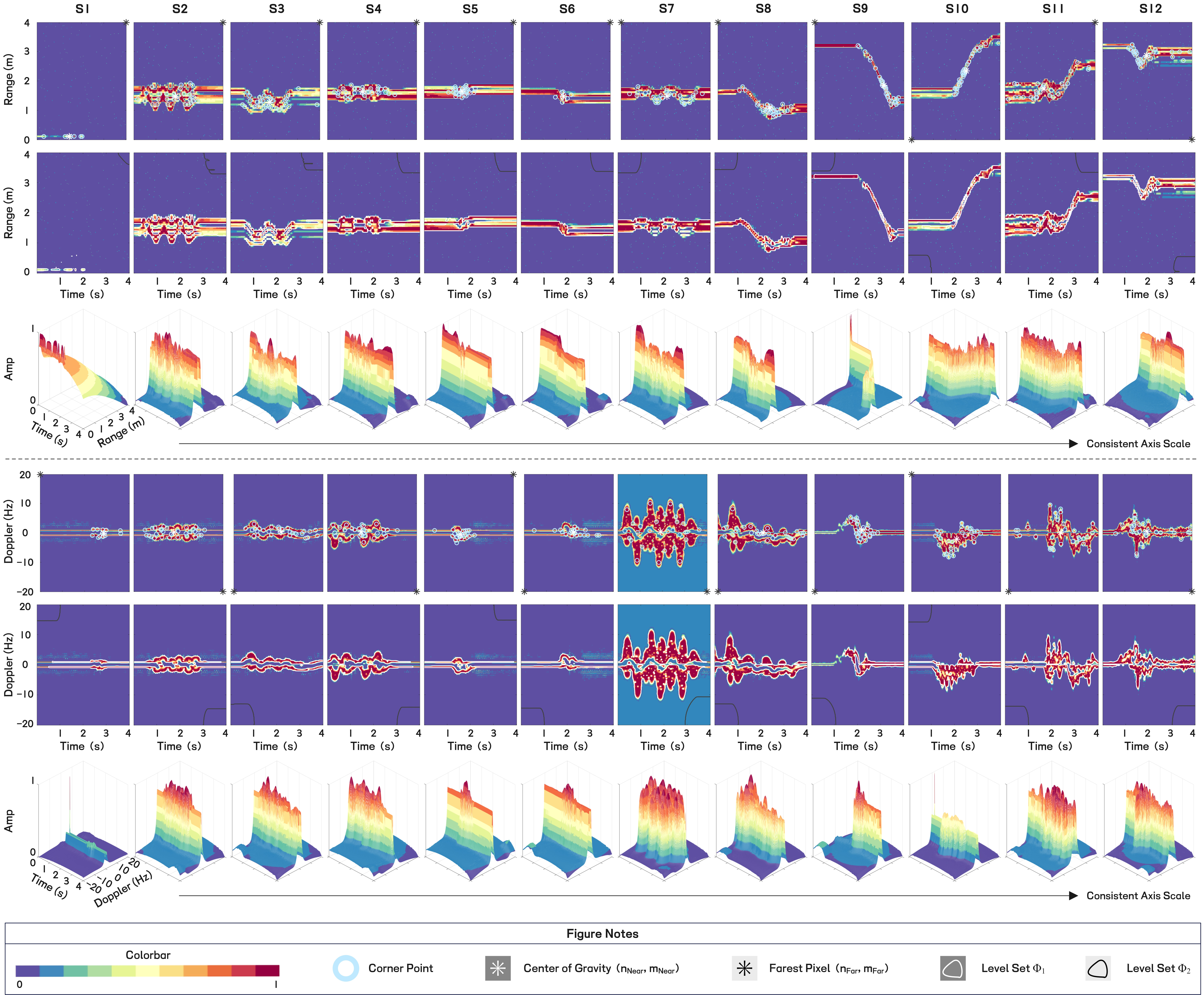}
    \caption{Simulated visualization results of the proposed method: The first row presents RTMs after corner detection, the second row presents RTMs after ACM-based feature extraction, the third row presents the extracted micro-Doppler signature on RTMs, the forth row presents DTMs after corner detection, the fifth row presents DTMs after ACM-based feature extraction, and the sixth row presents the extracted micro-Doppler signature on DTMs. $S1\sim S12$ are consistent with the predefined $12$ activity labels.}
    \label{Simulated Visualization}
    \vspace{-0.2cm}
\end{figure*}\par
\begin{figure*}[!ht]
    \centering
    \includegraphics[width=\textwidth]{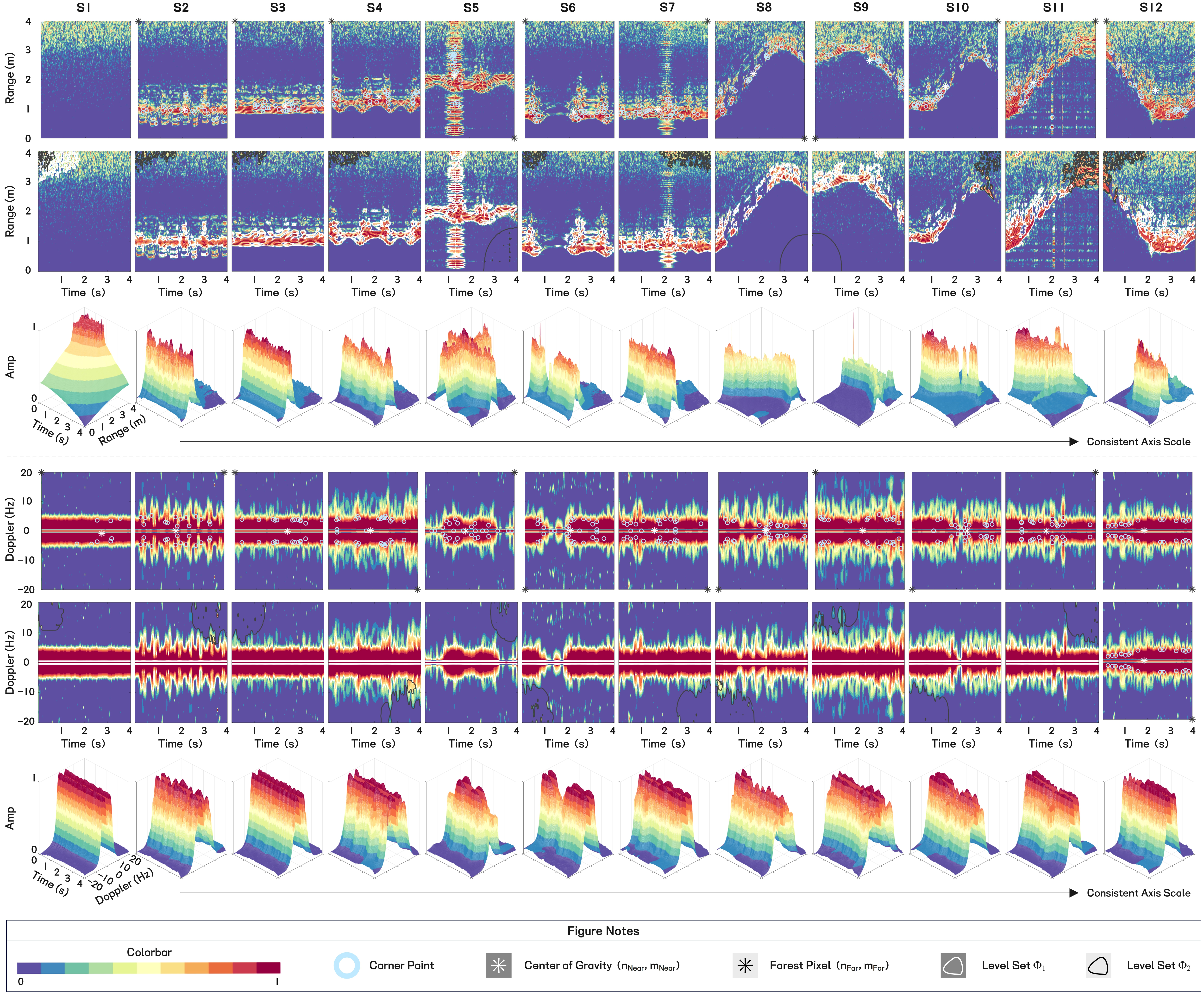}
    \caption{Measured visualization results of the proposed method: The first row presents RTMs after corner detection, the second row presents RTMs after ACM-based feature extraction, the third row presents the extracted micro-Doppler signature on RTMs, the forth row presents DTMs after corner detection, the fifth row presents DTMs after ACM-based feature extraction, and the sixth row presents the extracted micro-Doppler signature on DTMs. $S1\sim S12$ are consistent with the predefined $12$ activity labels.}
    \label{Measured Visualization}
    \vspace{-0.2cm}
\end{figure*}\par

\section{Numerical Simulations and Experiments}
In this section, numerical simulated and measured experiments demonstrate the effectiveness of the proposed method. First, the parameters and scene settings are introduced. Next, visualization experiments are presented. Then, experiments comparing recognition accuracy and robustness are analyzed. Next, ablation verifications are conducted. Finally, the experimental results are discussed.\par

\subsection{Parameter and Scene Settings}
The experiments in this paper use two sets of data, simulation and actual measurement, to verify the effectiveness of the proposed method. For the sake of rigor, most of the parameters and scene settings are kept consistent. The simulated data are generated numerically by combining the human motion capture data with the echo model from team UCL \cite{UCL}, and the measured data are collected from the built UWB TWR system in a typical urban building environment.\par
As shown in TABLE \ref{Parameter and Scene Settings}, consistent with the modeling section, a SISO TWR system is used to transmit and receive signals with the center frequency of $1.5\mathrm{~GHz}$ and the bandwidth of $2\mathrm{~GHz}$. The spacing between the transmitting and receiving antennas is $0.15~m$. Both the sampling points of the fast time dimension and slow time dimension are $1024$. The thickness of the wall is $0.12~m$ and the relative dielectric constant is around $6$. The wall in the simulation scenario is replaced with a rectangular homogeneous medium with the same parameters. The range of human motion is $1 \sim 4~m$ from radar with $12$ activities ($S1$, Empty; $S2$, Punching; $S3$, Kicking; $S4$, Grabbing; $S5$, Sitting Down; $S6$, Standing Up; $S7$, Rotating; $S8$, Walking; $S9$, Sitting to Walking; $S10$, Walking to Sitting; $S11$, Falling to Walking; $S12$, Walking to Falling) \cite{Gao3}. $4000$ sets of data are collected for both simulated and measured experiments, where $\frac{1}{5}$ of the data is used for performance verification of the proposed method. $20$ sets of data are randomly selected from each type of activity for template matching.\par
In order to achieve faster feature extraction and recognition speed with limited computational resources, all input images are resized to $256\times 256$ scale, which still meets the $7.5\mathrm{~cm}$ range resolution of TWR and the time resolution required for time-frequency analysis. The recommended settings for the hyperparameters at the current image scale are shown in Table \ref{Hyperparameter Settings}. It is recommended to dynamically adjust the hyper-parameter settings according to different data features, input image scales, and hardware resources.\par

\subsection{Visualization}
As shown in Fig. \ref{Simulated Visualization} and \ref{Measured Visualization}, both simulated and measured RTM and DTM images, the corner detection results, ACM-based level set functions, and micro-Doppler signature extraction results for $12$ types of activities are visualized.\par
From Fig. \ref{Simulated Visualization}, both simulated RTM and DTM are effective in labeling the corners at the critical moments of the human limb nodes. The centers of gravity of the corners all fall inside the curve. The level set $\Phi_1$ obtained by optimization with this point as the initiation can effectively focus the human motion micro-Doppler signature. The extracted micro-Doppler signature possesses the advantage of clear details and zero noise. From Fig. \ref{Measured Visualization}, similar conclusions can be obtained on RTMs. Unfortunately, the measured results show that the proposed method is sensitive to system interference. This will somewhat affect the accuracy of the subsequent recognition mapping. The feature extraction of the measured DTMs is poor. The key micro-Doppler information of some limb nodes is not effectively extracted after several rounds of evolution iterations. Therefore, although subsequent experiments will still compare, the proposed method is not recommended for recognition on measured DTMs.\par

\subsection{Comparative Experiments}
In this section, some existing network-based recognition methods are used to carry out comparative experiments, including four frontier image classification works: ResNet-50 \cite{TraditionalCNN}, VGG-19 \cite{TraditionalCNN}, ViT \cite{ViT}, and ConvNeXt \cite{ConvNeXt}. Also, the comparative methods include six frontier TWR HAR works: TWR-AEN-BiGRU \cite{Yang}, TWR-GCN \cite{Wang}, TWR-ResNeXt \cite{TWR-ResNeXt}, TWR-CapsuleNet \cite{TWR-CapsuleNet}, RPCA-ResNet \cite{RPCA-ResNet}, and TWR-WSN-CRF \cite{Gao2}. The hyperparameter settings and input image types of the comparative network methods are all consistent with \cite{Gao6}. The experimental results of recognition on RTM or DTM using the proposed method are compared separately.\par
\begin{figure}
    \centering
    \includegraphics[width=0.48\textwidth]{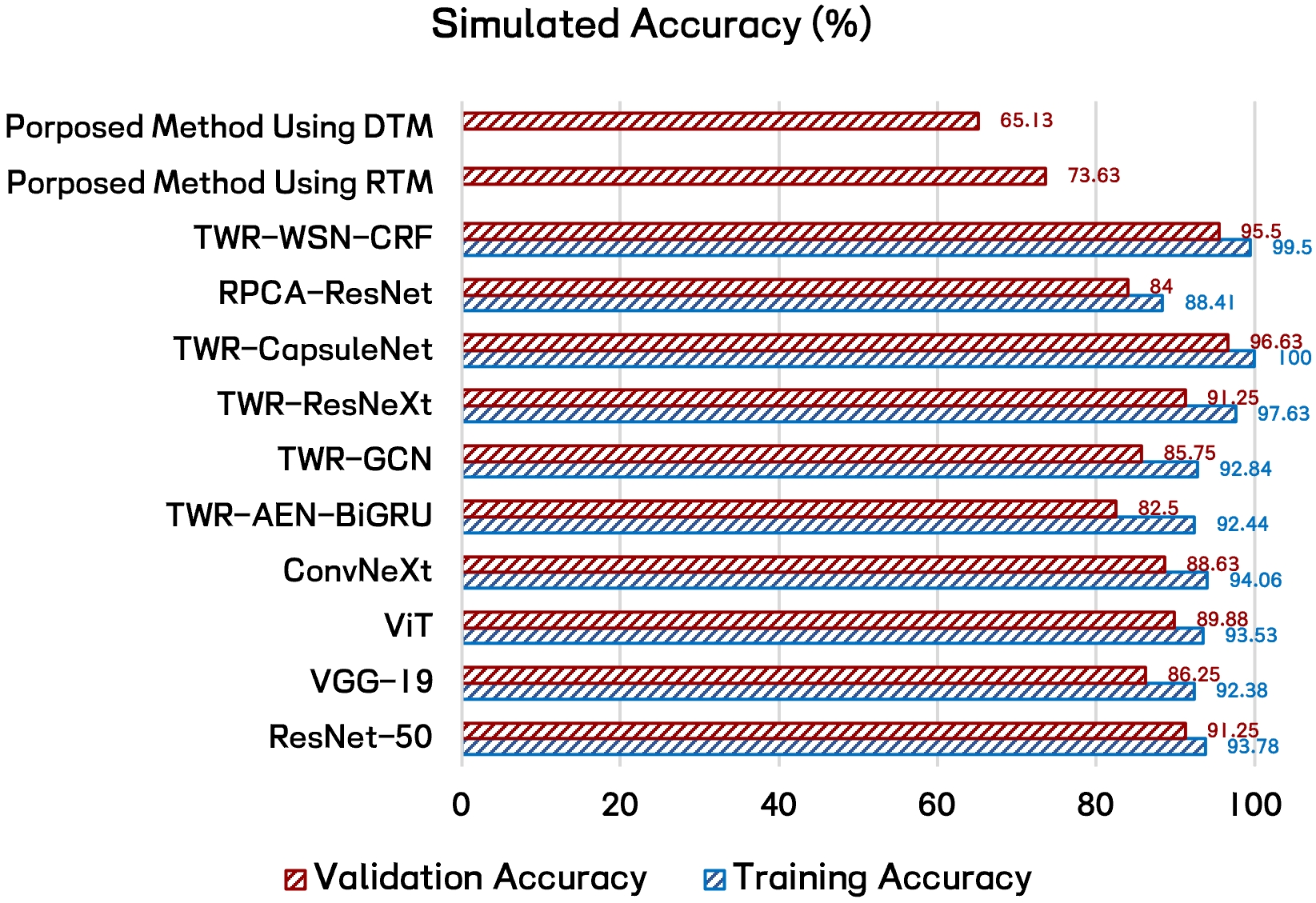}
    \caption{Simulated training and validation accuracy under different methods.}
    \label{Simulated Accuracy}
    \vspace{-0.0cm}
\end{figure}\par
\begin{figure}
    \centering
    \includegraphics[width=0.48\textwidth]{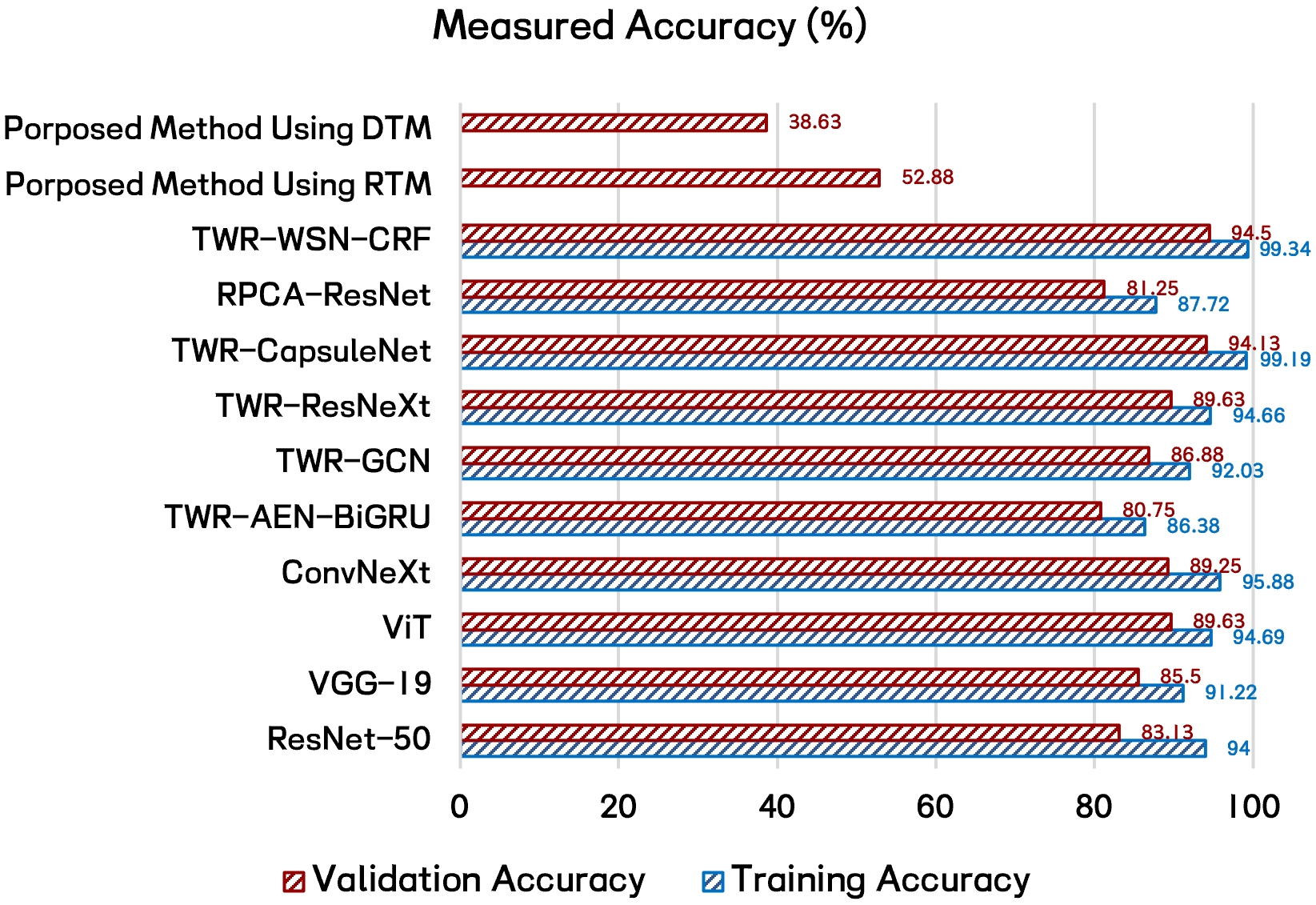}
    \caption{Measured training and validation accuracy under different methods.}
    \label{Measured Accuracy}
    \vspace{-0.2cm}
\end{figure}\par
\begin{table}[!ht]
\begin{center}
\caption{Simulated Robustness Testing$^{*}$.}\label{Simulated_Robustness}
\vspace{-0.0cm}
\resizebox{0.48\textwidth}{!}{
\begin{tabular}{c c c c c c c c}
\hline\hline
$\Delta$SNR (dB)$^{1}$ & $-12.00$ & $-10.00$ & $-8.00$ & $-6.00$ & $-4.00$ & $-2.00$ & $0.00$ \\
\hline
ResNet-50 & $68.75$ & $75.38$ & $80.00$ & $83.75$ & $86.63$ & $88.75$ & $91.25$ \\
VGG-19 & $62.13$ & $68.75$ & $73.75$ & $77.50$ & $80.13$ & $82.88$ & $86.25$ \\
ViT & $66.25$ & $72.50$ & $77.5$ & $81.25$ & $83.88$ & $86.25$ & $89.88$ \\
ConvNeXt & $65.38$ & $71.13$ & $76.63$ & $80.00$ & $82.50$ & $85.38$ & $88.63$ \\
TWR-AEN-BiGRU & $64.25$ & $73.88$ & $77.13$ & $79.00$ & $80.38$ & $81.13$ & $82.50$ \\
TWR-GCN & $75.38$ & $78.25$ & $80.13$ & $82.25$ & $84.00$ & $84.88$ & $85.75$ \\
TWR-ResNeXt & $77.38$ & $81.00$ & $84.88$ & $87.75$ & $89.38$ & $91.00$ & $91.25$ \\
TWR-CapsuleNet & $82.50$ & $88.50$ & $91.50$ & $93.75$ & $95.13$ & $95.88$ & $96.63$ \\
RPCA-ResNet & $61.13$ & $70.88$ & $76.50$ & $79.88$ & $81.25$ & $83.00$ & $84.00$ \\
TWR-WSN-CRF & $84.88$ & $87.38$ & $90.75$ & $92.00$ & $93.38$ & $93.75$ & $95.50$ \\
\textbf{Proposed / RTM} & $\mathbf{45.88}$ & $\mathbf{60.25}$ & $\mathbf{64.50}$ & $\mathbf{67.63}$ & $\mathbf{72.50}$ & $\mathbf{73.63}$ & $\mathbf{73.63}$ \\
\textbf{Proposed / DTM} & $\mathbf{39.00}$ & $\mathbf{50.25}$ & $\mathbf{53.50}$ & $\mathbf{56.75}$ & $\mathbf{61.75}$ & $\mathbf{64.00}$ & $\mathbf{65.13}$ \\
\hline \hline
\end{tabular}
}
\end{center}
\footnotesize $^{*}$ Validation accuracy ($\%$) of the proposed method under various SNR conditions. Comparative methods are consistent with Fig. \ref{Simulated Accuracy} and \ref{Measured Accuracy}.\\
\footnotesize $^{1}$ Decreased value of SNR (dB) after manually adding Gaussian noise with different variances to the echo.\\
\vspace{-0.0cm}
\end{table}\par
\begin{table}[!ht]
\begin{center}
\caption{Measured Robustness Testing$^{*}$.}\label{Measured_Robustness}
\vspace{-0.0cm}
\resizebox{0.48\textwidth}{!}{
\begin{tabular}{cccccccc}
\hline\hline
$\Delta$SNR (dB)$^{1}$ & $-12.00$ & $-10.00$ & $-8.00$ & $-6.00$ & $-4.00$ & $-2.00$ & $0.00$ \\
\hline
ResNet-50 & $67.63$ & $72.75$ & $75.88$ & $78.25$ & $80.25$ & $81.75$ & $83.13$ \\
VGG-19 & $69.63$ & $74.88$ & $78.13$ & $80.50$ & $82.63$ & $84.00$ & $85.50$ \\
ViT & $73.00$ & $78.50$ & $81.88$ & $84.38$ & $86.63$ & $88.13$ & $89.63$ \\
ConvNeXt & $72.63$ & $78.25$ & $81.50$ & $84.00$ & $86.25$ & $87.75$ & $89.25$ \\
TWR-AEN-BiGRU & $62.25$ & $69.88$ & $73.88$ & $77.00$ & $77.63$ & $79.50$ & $80.75$ \\
TWR-GCN & $72.25$ & $77.63$ & $80.63$ & $82.50$ & $83.75$ & $85.38$ & $86.88$ \\
TWR-ResNeXt & $74.88$ & $78.00$ & $80.50$ & $83.00$ & $86.50$ & $88.50$ & $89.63$ \\
TWR-CapsuleNet & $78.25$ & $83.13$ & $86.75$ & $89.88$ & $92.13$ & $93.38$ & $94.13$ \\
RPCA-ResNet & $55.75$ & $64.75$ & $69.38$ & $72.88$ & $76.63$ & $77.75$ & $81.25$ \\
TWR-WSN-CRF & $85.63$ & $88.50$ & $90.50$ & $91.13$ & $92.50$ & $93.50$ & $94.50$ \\
\textbf{Proposed / RTM} & $\mathbf{33.50}$ & $\mathbf{41.25}$ & $\mathbf{46.63}$ & $\mathbf{52.75}$ & $\mathbf{52.88}$ & $\mathbf{52.88}$ & $\mathbf{52.88}$ \\
\textbf{Proposed / DTM} & $\mathbf{23.63}$ & $\mathbf{30.00}$ & $\mathbf{31.88}$ & $\mathbf{35.75}$ & $\mathbf{35.75}$ & $\mathbf{38.63}$ & $\mathbf{38.63}$ \\
\hline \hline
\end{tabular}
}
\end{center}
\footnotesize $^{*}$ Validation accuracy ($\%$) of the proposed method under various SNR conditions. Comparative methods are consistent with Fig. \ref{Simulated Accuracy} and \ref{Measured Accuracy}.\\
\footnotesize $^{1}$ Decreased value of SNR (dB) after manually adding Gaussian noise with different variances to the echo.\\
\vspace{-0.2cm}
\end{table}\par
As shown in Fig. \ref{Simulated Accuracy} and \ref{Measured Accuracy}, the training accuracy and validation accuracy of the proposed method with existing methods are compared. Since the proposed method does not contain any network models, there is no concept of training accuracy. From Fig. \ref{Simulated Accuracy}, the simulated validation accuracy of existing methods is not less than $82\%$, and the validation accuracy of some methods is even more than $95\%$. The simulated validation accuracy of the proposed method on RTM and DTM is $73.63\%$ and $65.13\%$, respectively. This result has some gap relative to the network methods, but still has validity. From Fig. \ref{Measured Accuracy}, the measured validation accuracy of existing methods is not less than $80\%$. The simulated validation accuracy of the proposed method on RTM and DTM is $52.88\%$ and $38.63\%$, respectively. The proposed method still has some validity on the measured RTM. The results show that the proposed method is more suitable for RTM.\par
As shown in TABLE \ref{Simulated_Robustness} and \ref{Measured_Robustness}, the validation accuracy of the proposed method with the existing methods is compared under different SNR conditions. $\Delta$SNR denotes the decreased value of SNR in dB unit after manually adding Gaussian noise with different variances to the echo. As the SNR decreases, the less the accuracy of the method decreases, proving more robustness. From TABLE \ref{Simulated_Robustness}, the validation accuracy of the proposed method decreases by no more than $15\%$ when the SNR decreases by no more than $10\mathrm{~dB}$. From TABLE \ref{Measured_Robustness}, the validation accuracy of the proposed method decreases by no more than $12\%$ when the SNR decreases by no more than $10\mathrm{~dB}$. The results prove that the proposed method is consistent with or even better than the robustness of the vast majority of existing network methods.\par
\begin{table}[!ht]
\begin{center}
\caption{Ablation Experiment of Corner Detection$^{*}$.}\label{Corner_Ablation}
\vspace{-0.0cm}
\resizebox{0.48\textwidth}{!}{
\begin{tabular}{ccccc}
\hline\hline
\textbf{Method} & Harris \cite{Harris} & FAST \cite{FAST} & ECFRNet \cite{ECFRNet} & \textbf{Proposed} \\
\hline
Simulated RTM & $61.25$ & $72.88$ & $76.13$ & $\mathbf{73.63}$ \\
Simulated DTM & $52.63$ & $63.75$ & $68.25$ & $\mathbf{65.13}$ \\
Measured RTM & $40.00$ & $51.25$ & $50.75$ & $\mathbf{52.88}$ \\
Measured DTM & $26.25$ & $37.38$ & $38.38$ & $\mathbf{38.63}$ \\
\hline \hline
\end{tabular}
}
\end{center}
\footnotesize $^{*}$ Validation accuracy ($\%$) of the proposed method compared to three existing corner detection methods, where ECFRNet is a network-based method. The design of the other steps of the method is kept consistent with the theoretical section for the principle of control variables approach.\\
\vspace{-0.0cm}
\end{table}\par
\begin{table}[!ht]
\begin{center}
\caption{Ablation Experiment of ACM-Based Feature Extraction$^{*}$.}\label{ACM_Ablation}
\vspace{-0.0cm}
\resizebox{0.48\textwidth}{!}{
\begin{tabular}{ccccc}
\hline\hline
\textbf{Method} & LBF \cite{LBF} & GAC \cite{GAC} & DRLSE \cite{DRLSE} & \textbf{Proposed} \\
\hline
Simulated RTM & $42.25$ & $29.50$ & $58.00$ & $\mathbf{73.63}$ \\
Simulated DTM & $48.13$ & $31.75$ & $35.38$ & $\mathbf{65.13}$ \\
Measured RTM & $31.25$ & $25.88$ & $41.25$ & $\mathbf{52.88}$ \\
Measured DTM & $29.63$ & $17.50$ & $30.00$ & $\mathbf{38.63}$ \\
\hline \hline
\end{tabular}
}
\end{center}
\footnotesize $^{*}$ Validation accuracy ($\%$) of the proposed method compared to three existing ACM-based segmentation methods. The design of the other steps of the method is kept consistent with the theoretical section for the principle of control variables approach.\\
\vspace{-0.2cm}
\end{table}\par

\subsection{Ablation Verifications}
The proposed method consists of three main steps: Firstly, corner detection is achieved by SIFT. Then, feature extraction is achieved by multiphase Chan-Vese model. Finally, point cloud matching is achieved by Mapper algorithm. Method design for all three steps requires ablation verifications.\par
As shown in TABLE \ref{Corner_Ablation}, validation accuracy of the proposed method is compared to three existing corner detection methods, where Harris \cite{Harris} and FAST \cite{FAST} are traditional machine-learning-based corner detection method and ECFRNet \cite{ECFRNet} is neural-network-based corner detection method. The design of the other steps of the method is kept consistent with the theoretical section. From simulated RTM and DTM results, for SIFT, ECFRNet and FAST, which possess good image noise robustness, the final validation accuracy does not vary much. This demonstrates that the center of gravity $(n_\mathrm{Near},m_\mathrm{Near})$ as well as the farthest point $(n_\mathrm{Far},m_\mathrm{Far})$ can be projected inside the curve, effectively initiating subsequent feature extraction. Harris method is sensitive to noise and performs worse with more errors in detecting corners. Similar conclusions can be drawn from measured RTM and DTM results. The above findings together prove the rationality of the design of corner detection method.\par
As shown in TABLE \ref{ACM_Ablation}, validation accuracy of the proposed method is compared to three existing ACM-based feature extraction methods, including LBF \cite{LBF}, GAC \cite{GAC}, and DRLSE \cite{DRLSE}. The design of the other steps of the method is kept consistent with the theoretical section. From simulated RTM and DTM results, except for utilizing the proposed multiphase Chan-Vese model, the other three methods all perform poorly in terms of validation accuracy. With the exception of DRLSE on simulated RTMs, none of the other methods are able to recognize data that is more than half as accurate. Similar conclusions can be drawn from measured RTM and DTM results. None of the comparative methods can exceed $42\%$ accuracy on measured data. The above findings together prove the rationality of the design of ACM-based micro-Doppler signature extraction method.\par
\begin{table}[!ht]
\begin{center}
\caption{Ablation Experiment of Point Cloud Matching$^{*}$.}\label{PointCloud_Ablation}
\vspace{-0.0cm}
\resizebox{0.48\textwidth}{!}{
\begin{tabular}{cccc}
\hline\hline
\textbf{Method} & Hausdorff \cite{Hausdorff} & Wasserstein \cite{Wasserstein} & \textbf{Proposed} \\
\hline
Simulated RTM & $55.88$ & $58.25$ & $\mathbf{73.63}$ \\
Simulated DTM & $45.75$ & $59.63$ & $\mathbf{65.13}$ \\
Measured RTM & $39.25$ & $37.63$ & $\mathbf{52.88}$ \\
Measured DTM & $30.13$ & $28.50$ & $\mathbf{38.63}$ \\
\hline \hline
\end{tabular}
}
\end{center}
\footnotesize $^{*}$ Validation accuracy ($\%$) of the proposed method compared to two existing metrics of measuring point cloud similarity. The design of the other steps of the method is kept consistent with the theoretical section for the principle of control variables approach.\\
\vspace{-0.2cm}
\end{table}\par
As shown in TABLE \ref{PointCloud_Ablation}, validation accuracy of the proposed method is compared to two existing metrics of measuring point cloud similarity, including Hausdorff distance \cite{Hausdorff} and Wasserstein distance \cite{Wasserstein}. The HAR is achieved by finding the smallest category of total distance between the input point cloud and the template point clouds. The design of the other steps of the method is kept consistent with the theoretical section. From simulated RTM and DTM results, estimating point cloud similarity using Mapper's algorithm is better than directly using distance metrics. The effect of this enhancement is even more pronounced on measured RTMs and DTMs, where a $10\%$ gain in validation accuracy or even more can be achieved. The above findings together prove the rationality of the design of point cloud matching-based HAR method.\par

\subsection{Discussion}
Through the above visualization, accuracy comparison, robustness comparison, and ablation validation of each step of the proposed method, the effectiveness of the proposed method is proved. but also found many limitations. However, the results also revealed numerous limitations of the method design, including:\par
\textbf{(1) Limitations of the Overall Logic:} Once again, it is important to emphasize that this work is forcibly designed to eschew network models for achieving intelligent recognition tasks. This not only reduces the validation accuracy, but is also limited by the design of the multi-stage optimization algorithm, which is costly in terms of inference time. If it is necessary to summarize one advantage of this work, it would be that the need for scenario prior data is drastically reduced.\par
\textbf{(2) Limitations of the Micro-Doppler Signature Extraction Method:} The multiphase Chan-Vese model is sensitive to TWR system interference when extracting micro-Doppler signature. Interference signal is incorrectly extracted as micro-Doppler signature. There is a need to develop ACM methods that are specifically applicable to radar images.\par
\textbf{(3) Limitations of the HAR Method:} Based on the point cloud features, the template matching method using the collected data is certainly effective, but if it can be combined with the indoor human motion model to directly achieve the complex activity recognition, the method might be more underlying feasibility.\par

\section{Conclusion}
This paper has proposed to return to traditional ideas by avoiding neural networks for the task of TWR HAR, with the aim of achieving intelligent recognition as well as the network models. In detail, the RTM and DTM of TWR have first been generated. Then, the initial regions of the human target foreground and noise background on the maps have been determined using the corner detection method, and the micro-Doppler signature has been segmented using the multiphase ACM method. The micro-Doppler segmentation feature has been discretized into a two-dimensional point cloud. Finally, the topological similarity between the resulting point cloud and the point clouds of the template data has been calculated using the Mapper algorithm to obtain the recognition results. The effectiveness of the proposed method has been demonstrated through numerical simulations and measured experiments.\par

\section*{Acknowledgement}
This paper is dedicated to the memory of my grandma.\par
\begin{figure}[htbp]
    \centering
    \includegraphics[width=0.48\textwidth]{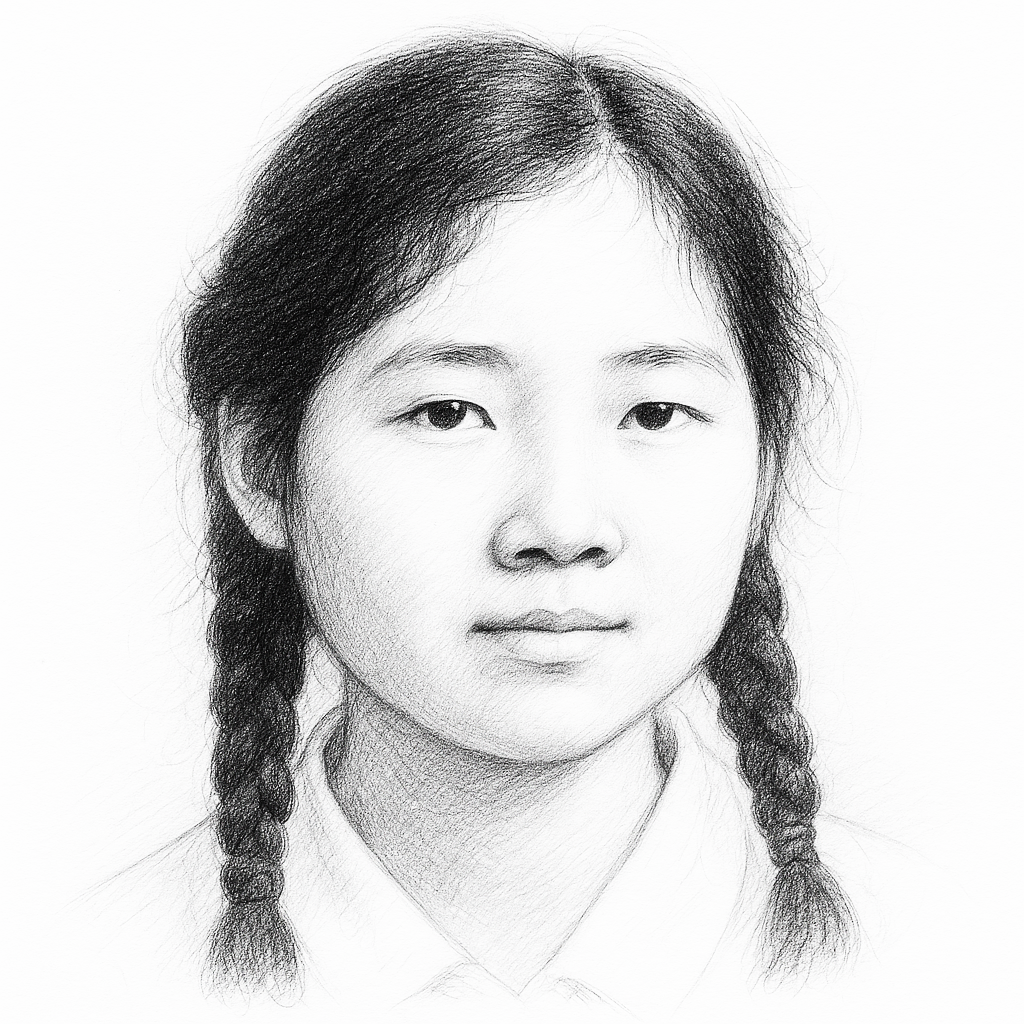}
    \caption{A sketch of my grandma in her youth.}
    \label{Grandma}
    \vspace{-0.0cm}
\end{figure}\par
I would like to thank my family: My grandfather, mother, father, aunt, for the impeccable care you gave to my grandmother on her deathbed and for making me feel the most precious affection on earth. Thank you to all the elders and friends who have been cared for and concerned with my grandmother during this time, you are all benefactors to our family.\par
Thank you to my mentors for nurturing and trusting me with the ability to do this unique work. Additionally, thanks to my love and my fellow close friend, it is our commitment that makes this work possible.\par
I'll do my best to pursue scientific research till death, but such a work will probably only come this once in my life. I truly hope everyone can find something in my journey. It is the power of love that penetrates all difficulties.\par


\end{document}